\documentclass{article}

\usepackage[nonatbib,final]{include/neurips_2021}
\usepackage[numbers]{natbib}

\usepackage[utf8]{inputenc} %
\usepackage[T1]{fontenc}    %
\usepackage{babel}
\usepackage{textcomp}
\usepackage{url}            %
\usepackage{booktabs}       %
\usepackage{amsfonts}       %
\usepackage{nicefrac}       %
\usepackage{microtype}      %
\usepackage{graphicx}
\usepackage{subfigure}
\usepackage{xcolor}
\usepackage{bm}
\usepackage{tabularx}
\usepackage{wrapfig}
\usepackage[ruled,vlined]{algorithm2e}
\usepackage{multicol}
\usepackage{rotating}
\usepackage{hyperref}
\hypersetup{
    colorlinks=true,
    linkcolor=magenta,
    citecolor=blue,
    urlcolor=blue,
}
     
\let\oldhref\href
\renewcommand{\href}[2]{\oldhref{#1}{\texttt{\bfseries#2}}}

\newcommand{\email}[2][black]{\href{mailto:#2}{\color{#1}\,#2\,}}

\newcommand{\algoabbrev}{REPAIRED}

\newcommand{\plrabbrev}{$\textnormal{PLR}^{\perp}$}

\definecolor{dark-gray}{gray}{0.35}

\newcounter{commentCounter}
\newif\iftrvar
\trvarfalse
\iftrvar
\newcommand{\tim}[1]{{\small \color{red} \refstepcounter{commentCounter}\textsf{[TR]$_{\arabic{commentCounter}}$:{#1}}}}
\newcommand{\minqi}[1]{{\small \color{blue} \refstepcounter{commentCounter}\textsf{[MJ]$_{\arabic{commentCounter}}$:{#1}}}}
\newcommand{\ed}[1]{{\small \color{magenta} \refstepcounter{commentCounter}\textsf{[ETG]$_{\arabic{commentCounter}}$:{#1}}}}

\newcommand{\jcom}[1]{\textcolor{purple}{JPH: #1}}

\newcommand{\michael}[1]{{\small \color{cyan} \refstepcounter{commentCounter}\textsf{[MDD]$_{\arabic{commentCounter}}$:{#1}}}}

\newcommand{\jakob}[1]{{\small \color{green} \refstepcounter{commentCounter}\textsf{[JF]$_{\arabic{commentCounter}}$:{#1}}}}

\else
\newcommand{\tim}[1]{}
\newcommand{\ed}[1]{}
\newcommand{\minqi}[1]{}
\newcommand{\jcom}[1]{}
\newcommand{\michael}[1]{}
\newcommand{\jakob}[1]{}

\fi

\usepackage{amsmath}
\usepackage{amsthm}
\usepackage{stmaryrd}

        {\medskip}

\usepackage{amsthm}

\newtheorem{innercustomgeneric}{\customgenericname}
\providecommand{\customgenericname}{}
\newcommand{\newcustomtheorem}[2]{%
  \newenvironment{#1}[1]
  {%
   \renewcommand\customgenericname{#2}%
   \renewcommand\theinnercustomgeneric{##1}%
   \innercustomgeneric
  }
  {\endinnercustomgeneric}
}

\newcustomtheorem{customthm}{Theorem}

        {\hspace*{\fill}$\Box$\par\vspace{4mm}}
        {\hspace*{\fill}$\Box$\par}
        
\DeclareMathOperator*{\argmax}{arg\!\max}
\DeclareMathOperator*{\argmin}{arg\!\min}
\newcommand{\EO}{\mathop{\mathbb{E}}}

\newcommand{\Specialize}[2]{{#1}^{#2}}

\newcommand{\PPOMDP}{\mathcal{M}}

\newcommand{\As}{A}
\newcommand{\Os}{O}
\newcommand{\Ss}[1]{\Specialize{S}{#1}}
\newcommand{\Tf}[1]{\Specialize{\mathcal{T}}{#1}}
\newcommand{\Of}[1]{\Specialize{\mathcal{I}}{#1}}
\newcommand{\Rf}[1]{\Specialize{\mathcal{R}}{#1}}
\newcommand{\discount}{\gamma}

\newcommand{\Ns}{\Theta}
\newcommand{\apply}[2]{#1_{#2}}
\newcommand{\Dist}[1]{\mathbf{\Delta}(#1)}

\usepackage{cleveref}

\title{Replay-Guided Adversarial Environment Design}

\author{%
    Minqi Jiang\thanks{Equal contribution. Correspondence to \email{msj@fb.com} and \email{michael\_dennis@berkeley.edu}.} \\UCL, FAIR 
    \And Michael Dennis\footnotemark[1] \\UC Berkeley
    \And Jack Parker-Holder \\ University of Oxford
    \AND Jakob Foerster \\ FAIR
    \And Edward Grefenstette  \\ UCL, FAIR
    \And Tim Rockt\"{a}schel \\ UCL, FAIR
}

\begin{document}

\maketitle

\begin{abstract}
Deep reinforcement learning (RL) agents may successfully generalize to new settings if trained on an appropriately diverse set of environment and task configurations. Unsupervised Environment Design (UED) is a promising self-supervised RL paradigm, wherein the free parameters of an underspecified environment are automatically adapted during training to the agent's capabilities, leading to the emergence of diverse training environments. Here, we cast Prioritized Level Replay (PLR), an empirically successful but theoretically unmotivated method that selectively samples randomly-generated training levels, as UED. We argue that by curating completely random levels, PLR, too, can generate novel and complex levels for effective training. This insight reveals a natural class of UED methods we call Dual Curriculum Design (DCD). 
Crucially, DCD includes both PLR and a popular UED algorithm, PAIRED, as special cases and inherits similar theoretical guarantees.  
This connection allows us to develop novel theory for PLR, providing a version with a robustness guarantee at Nash equilibria. 
Furthermore, our theory suggests a highly counterintuitive improvement to PLR: by stopping the agent from updating its policy on uncurated levels (training on \emph{less} data), we can improve the convergence to Nash equilibria. 
Indeed, our experiments confirm that our new method, \plrabbrev{}, obtains better results on a suite of out-of-distribution, zero-shot transfer tasks, in addition to demonstrating that \plrabbrev{} improves the performance of PAIRED, from which it inherited its theoretical framework.
\end{abstract}

\section{Introduction}
While deep reinforcement learning (RL) approaches have led to many successful applications in challenging domains like Atari~\citep{mnih2015human}, Go~\citep{silver2016mastering}, Chess~\citep{silver2018general}, Dota~\citep{berner2019dota}, and StarCraft~\citep{vinyals2019grandmaster} in recent years, deep RL agents still prove to be brittle, often failing to transfer to environments only slightly different from those encountered during training~\citep{zhang2018dissection,coinrun}. To ensure learning of robust and well-generalizing policies, agents must train on sufficiently diverse and informative variations of environments (e.g.~see Section 3.1 of \citep{procgen_benchmark}). 
However, it is not always feasible to specify an appropriate training distribution or a generator thereof. Agents may therefore benefit greatly from methods that automatically adapt the distribution over environment variations throughout training~\citep{paired,plr}. Throughout this paper we will call a particular environment instance or configuration (e.g.~an arrangement of blocks, race tracks, or generally any of the environment's constituent entities) a \emph{level}. 

Two recent works~\citep{paired,plr} have sought to empirically demonstrate this need for a more targeted agent-adaptive mechanism for selecting levels on which to train RL agents, so to ensure efficient learning and generalization to unseen levels---as well as to provide methods implementing such mechanisms. 
The first method, Protagonist Antagonist Induced Regret Environment Design (PAIRED)~\citep{paired}, introduces a self-supervised RL paradigm called Unsupervised Environment Design (UED).
Here, an environment generator (a \emph{teacher}) is co-evolved with a \emph{student} policy that trains on levels actively proposed by the teacher, leading to a form of adaptive curriculum learning. The aim of this coevolution is for the teacher to gradually learn to generate environments that exemplify properties of those that might be encountered at deployment time, and for the student to simultaneously learn a good policy that enables zero-shot transfer to such environments. PAIRED's specific adversarial approach to environment design ensures a useful robustness characterization of the final student policy in the form of a minimax regret guarantee \citep{savage1951theory}---assuming that its underlying teacher-student multi-agent system arrives at a Nash equilibrium~\cite[NE,][]{nash1950equilibrium}. 
In contrast, the second method, Prioritized Level Replay (PLR)~\citep{plr}, embodies an alternative form of dynamic curriculum learning that does not assume control of level generation, but instead, the ability to selectively replay existing levels. PLR tracks levels previously proposed by a black-box environment generator, and for each, estimates the agent's learning potential in that level, in terms of how useful it would be to gather new experience from that level again in the future. The PLR algorithm exploits these scores to adapt a schedule for revisiting or \emph{replaying} levels to maximize learning potential. PLR has been shown to produce scalable and robust results, improving both sample complexity of agent training and the generalization of the learned policy in diverse environments. However, unlike PAIRED, PLR is motivated with heuristic arguments and lacks a useful theoretical characterization of its learning behavior.

In this paper, we argue that PLR is, in and of itself, an effective form of UED: Through curating even randomly generated levels, PLR can generate novel and complex levels for learning robust policies. This insight leads to a natural class of UED methods which we call \emph{Dual Curriculum Design} (DCD). In DCD, a student policy is challenged by a team of two co-evolving teachers. One teacher actively generates new, challenging levels, while the other passively curates existing levels for replaying, by prioritizing those estimated to be most suitably challenging for the student. We show that PAIRED and PLR are distinct members of the DCD class of algorithms and prove in Section~\ref{sec:dual} that all DCD algorithms enjoy similar minimax regret guarantees to that of PAIRED.
 
We make use of this result to provide the first theoretical characterization of PLR, which immediately suggests a simple yet highly counterintuitive adjustment to PLR: By only training on trajectories in replay levels, PLR becomes provably robust at NE. We call this resulting variant \plrabbrev{} (Section~\ref{sec:robust_plr}). From this perspective, PLR effectively performs level design in a diametrically opposite manner to PAIRED---through prioritized selection rather than active generation. A second corollary to the provable robustness of DCD algorithms shows that \plrabbrev{} can be extended to make use of the PAIRED teacher as a level generator while preserving the robustness guarantee of PAIRED, resulting in a method we call \emph{Replay-Enhanced PAIRED} (REPAIRED) (Section~\ref{sec:repaired}). We hypothesize that in this arrangement, \plrabbrev{} plays a complementary role to PAIRED in robustifying student policies.

Our experiments in Section~\ref{sec:experiments}  investigate the learning dynamics of \plrabbrev{}, REPAIRED, and their replay-free counterparts on a challenging maze domain and a novel continuous control UED setting based on the popular CarRacing environment \citep{gym}. In both of these highly distinct settings, our methods provide significant improvements over PLR and PAIRED,  producing agents that can perform out-of-distribution (OOD) generalization to a variety of human designed mazes and Formula 1 tracks.

\begin{figure}[t!]
    \centering
    \begin{minipage}{0.99\textwidth}
    \centering\subfigure[DR]{\includegraphics[width=.18\linewidth]{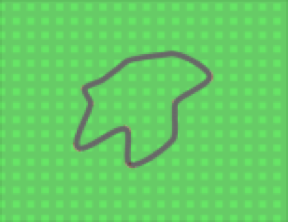}}    
    \centering\subfigure[PAIRED]{\includegraphics[width=.18\linewidth]{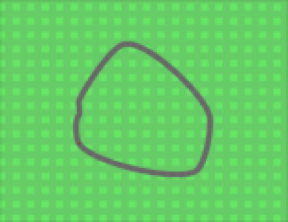}}
    \centering\subfigure[REPAIRED]{\includegraphics[width=.18\linewidth]{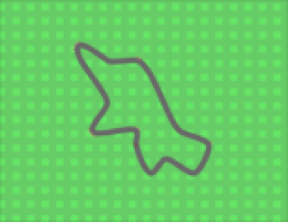}}
    \centering\subfigure[PLR$^{\bot}$]{\includegraphics[width=.18\linewidth]{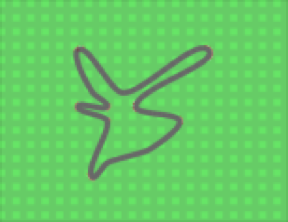}}
    \centering\subfigure[Human]{\includegraphics[width=.18\linewidth]{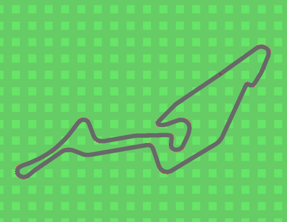}} 
    \caption{\small{Randomly drawn samples of CarRacing tracks produced by different methods. (a)~Domain Randomization (DR) produces tracks of average complexity, with few sharp turns. (b)~PAIRED often overexploits the difference in the students, leading to simple tracks that incidentally favor the antagonist. (c)~REPAIRED mitigates this degeneracy, recovering track complexity. (d)~\plrabbrev{} selects the most challenging randomly generated tracks, resulting in tracks that more closely resemble human-designed tracks, such as  (e)~the N\"{u}rburgring Grand Prix.
    }}
    \label{figure:carracing_pics}
    \end{minipage}
    \vspace{-.5cm}
\end{figure}

In summary, we present the following contributions:
(i) We establish a common framework, Dual Curriculum Design, that encompasses PLR and PAIRED. This allows us to develop new theory, which provides the first robustness guarantees for PLR at NE as well as for REPAIRED, which augments PAIRED with a PLR-based replay mechanism.
(ii) Crucially, our theory suggests a highly counterintuitive improvement to PLR: the convergence to NE should be assisted by training on less data when using PLR---namely by only taking gradient updates from data that originates from the PLR buffer, using the samples from the environment distribution only for computing the prioritization of levels in the buffer.
(iii) Our experiments in a maze domain and a novel car-racing domain show that our methods significantly outperform their replay-free counterparts in zero-shot generalization. We open source our methods at 
\href{https://github.com/facebookresearch/dcd}{https://github.com/facebookresearch/dcd}.

\section{Background}
\label{sec:background}

\subsection{Unsupervised Environment Design}
\label{ssec:ued}

Unsupervised Environment Design (UED), as introduced by~\citep{paired}, is the problem of automatically designing a distribution of environments that adapts to the learning agent.
UED is defined in terms of an \emph{Underspecified POMDP} (UPOMDP), given by $\PPOMDP = \langle \As,\Os, \Ns, \Ss{\PPOMDP}, \Tf{\PPOMDP},\Of{\PPOMDP},\Rf{\PPOMDP},\discount \rangle$, where $\As$ is a set of actions, $\Os$ is a set of observations, $\Ss{}$ is a set of states, $\Tf{}:  S \times A \times \Ns \rightarrow \Dist{S}$ is a transition function, $\Of{}: S \rightarrow O$ is an observation (or inspection) function, $\Rf{}: S \rightarrow \mathbb{R}$ is a reward function, and $\discount$ is a discount factor. This definition is identical to a POMDP with the addition of $\Theta$ to represent the free-parameters of the environment. These parameters can be distinct at every time step and incorporated into the transition function $\Tf{\PPOMDP}: S \times A \times \Ns \rightarrow \Dist{S}$.  
For example, $\Theta$ could represent the possible positions of obstacles in a maze. We will refer to the environment resulting from a fixed $\theta \in \Theta$ as $\apply{\PPOMDP}{\theta}$, or with a slight abuse of notation, simply $\theta$ when clear from context. We define the value of $\pi$ in $\apply{\PPOMDP}{\theta}$ to be
$V^{\theta}(\pi) = \EO[\sum_{i=0}^{T} r_t\gamma^t]$ where $r_t$ are the rewards attained by $\pi$ in $\apply{\PPOMDP}{\theta}$.  Aligning with terminology from~\citep{plr}, we refer to a fully-specified environment as a \emph{level}.

\subsection{Protagonist Antagonist Induced Regret Environment Design}
\label{ssec:paired}

Protagonist Antagonist Induced Regret Environment Design~\cite[PAIRED,][]{paired} presents a UED approach consisting of simultaneously training agents in a three player game: the protagonist $\pi_A$ and the antagonist $\pi_B$ are trained in environments generated by the teacher $\tilde{\theta}$. The objective of this game is defined by
$U(\pi_A,\pi_B,\tilde{\theta}) = \EO_{\theta \sim \tilde{\theta}} [\normalfont\textsc{Regret}^{\theta}(\pi_A,\pi_B)]$,
where regret is defined by $\normalfont\textsc{Regret}^{\theta}(\pi_A,\pi_B)= V^{\theta}(\pi_B) - V^{\theta}(\pi_A)$. The protagonist and antagonist are both trained to maximize their discounted environment returns while the teacher is trained to maximize $U$. 
Note that by maximizing regret, the teacher is disincentivized from generating unsolvable levels, which will have a maximum regret of $0$.
As shorthand, we will sometimes refer to the protagonist and antagonist jointly as the \emph{student agents}. The counterclockwise loop beginning at the student agents in Figure~\ref{figure:repaired_overview} summarizes this approach, with the students being both the protagonist and antagonist.

As both student agents grow more adept at solving different levels, the teacher continues to adapt its level designs to exploit the weaknesses of the protagonist in relation to the antagonist. As this dynamic unfolds, PAIRED produces an emergent curriculum of progressively more complex levels along the boundary of the protagonist's capabilities. PAIRED is a creative method in the sense that the teacher may potentially generate an endless sequence of novel levels.
However, as the teacher only adapts through gradient updates, it is inherently slow to adapt to changes in the student policies. 

\subsection{Prioritized Level Replay}
Prioritized Level Replay~\cite[PLR,][]{plr} is an active-learning strategy shown to improve a policy's sample efficiency and generalization to unseen levels when training and evaluating on levels from a common UPOMDP, typically implemented as a seeded simulator. PLR maintains a level buffer $\bm{\Lambda}$ of the top $K$ visited levels with highest learning potential as estimated by the time-averaged L1 value loss of the learning agent over the last episode on each level. At the start of each training episode, with some predefined replay probability $p$, PLR uses a bandit to sample the level from $\bm{\Lambda}$ to maximize the estimated learning potential; otherwise, with probability $1-p$, PLR samples a new level from the simulator. 
In contrast to the generative but slow-adapting PAIRED, PLR does not create new levels, but instead, acts as a fast-adapting curation mechanism for selecting the next training level among previously encountered levels. Also unlike PAIRED, PLR does not provide a robustness guarantee. By extending the theoretical foundation of PAIRED to PLR, we will show how PLR can be modified to provide a robustness guarantee at NE, as well as how PAIRED can exploit PLR's complementary curation to quickly switch among generated levels to maximize the student's regret.

\section{The Robustness of Dual Curriculum Design}
\label{sec:dual}
\begin{figure}[h]
\vspace{-3mm}
    \centering{\includegraphics[width=1\linewidth]{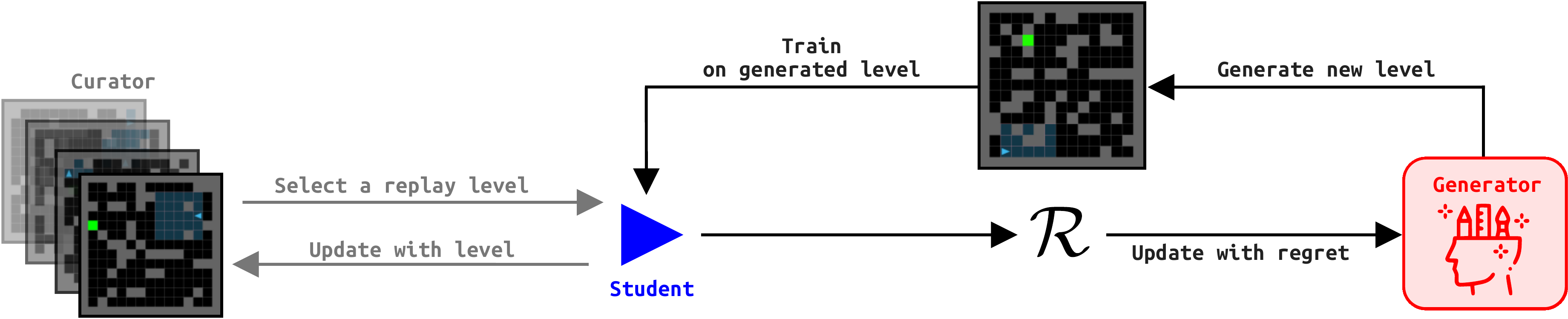}}
    \caption{\small{Overview of Dual Curriculum Design (DCD). The student learns in the presence of two co-adapting teachers that aim to maximize the student's regret: The generator teacher designs new levels to challenge the agent, and the curator teacher prioritizes a set of levels already created, selectively sampling them for replay.}} 
    \label{figure:repaired_overview}
\end{figure}

The previous approaches of PAIRED and PLR reveal a natural duality: Approaches that gradually learn to generate  levels like PAIRED, and methods which cannot generate levels, but instead, quickly curate existing ones, like PLR. This duality suggests combining slow level generators with fast level curators. We call this novel class of UED algorithms Dual Curriculum Design (DCD). For instance, PLR can be seen as curator with a prioritized sampling mechanism with a random generator, while PAIRED, as a regret-maximizing generator without a curator.
DCD can further consider Domain Randomization (DR) as a degenerate case of a random level generator without a curator.

To theoretically analyze this space of methods, we model DCD as a three player game among a student agent and two teachers called the \emph{dual curriculum game}. However, to formalize this game, we must first formalize the single-teacher setting: Suppose the UPOMDP is clear from context. Then, given a utility function for a single teacher, $U_t(\pi,\theta)$, we can naturally define the \emph{base game} between the student $s$ and teacher $t$ as $G = \langle S = S_s \times S_t, U = U_s \times U_t \rangle$, where $S_s= \Pi$ is the strategy set of the student, $S_t = \Theta$ is the strategy set of the teacher, and $U_s(\pi, \theta) = V^\theta(\pi)$ is the utility function of the student.  In Sections \ref{sec:robust_plr} and \ref{sec:repaired}, we will study settings corresponding to different choices of utility functions for the teacher agents, namely the maximum-regret objective $U_t^R(\pi, \theta)$ and the uniform objective $U_t^U(\pi, \theta)$. These two objectives are defined as follows (for any constant $C$):

\begin{minipage}[t]{\textwidth}
\vspace{-.3cm}

\begin{minipage}[t]{1\textwidth} 
  \begin{equation}
  U_t^R(\pi, \theta) = \argmax_{\pi^* \in \Pi}\{V^\theta(\pi^*)-V^\theta(\pi)\}
  \end{equation} 
\end{minipage} 
\begin{minipage}[t]{1\textwidth} 
  \begin{equation}
  U_t^U(\pi, \theta) = C
  \end{equation}
\end{minipage}
\end{minipage}

In the dual curriculum game $\overline{G}$, the first teacher plays the game with probability $p$, and the second, with probability $(1-p)$---or more formally, $\overline{G} = \langle \overline{S} = S_s \times S_t \times S_t, U = \overline{U}_s \times \overline{U}^1_t \times \overline{U}^2_t\rangle$, where the utility functions for the student and two teachers respectively, $\overline{U}_s, \overline{U}^1_t, \overline{U}^2_t$, are defined as follows:

\begin{minipage}[t]{\textwidth}
\vspace{-.2cm}

\begin{minipage}[t]{1\textwidth} 
  \begin{equation}
  \overline{U}^1_t(\pi, \theta^1, \theta^2) = p U_t^1(\pi, \theta^1)
  \end{equation}
\end{minipage}
\begin{minipage}[t]{1\textwidth} 
  \begin{equation}
  \overline{U}^2_t(\pi, \theta^1, \theta^2)   = (1-p)U_t^2(\pi, \theta^2)
  \end{equation} 
\end{minipage}

\begin{minipage}[t]{1\textwidth} 
  \begin{equation}
  \overline{U}_s(\pi, \theta^1, \theta^2)  = p U_s(\pi, \theta^1) + (1-p)U_s(\pi, \theta^2)
  \end{equation} 
\end{minipage} 
\vspace{0.2cm}

\end{minipage}

Our main theorem is that NE in the dual curriculum game are approximate NE of both the base game for either of the original teachers and the base game with a teacher maximizing the joint-reward of $pU^1_t + (1-p)U^2_t$, where the quality of the approximations depends on the mixing probability $p$.

\vspace{0.1cm}
\newtheorem{main_theorem}{Theorem}
\begin{main_theorem}
\label{theorem:main}
Let $B$ be the maximum difference between $U^1_t$ and $U^2_t$, and let $(\pi, \theta^1, \theta^2)$ be a NE for $\overline{G}$.  Then $(\pi, p\theta^1+(1-p)\theta^2)$ is an approximate NE for the base game with either teacher or for a teacher optimizing their joint objective. More precisely, it is a $2Bp(1-p)$-approximate NE when $U_t = pU^1_t+(1-p)U^2_t$, a $2B(1-p)$-approximate NE when $U_t = U^1_t$, and a $2Bp$-approximate NE when $U_t = U^2_t$.
\end{main_theorem}

The intuition behind this theorem is that, since the two teachers do not affect each other's behavior, their best response to a fixed $\pi_s$ is to choose a strategy $\theta$ that maximizes $U_t^1$ and $U_t^2$ respectively. Moreover, the two teachers' strategies can be viewed as a single combined strategy for the base game with the joint-objective, or with each teacher's own objective. In fact, the teachers provide an approximate best-response to each case of the base game simply by playing their individual best responses. Thus, when we reach a NE of the dual curriculum game, the teachers arrive at approximate best responses for both the base game with the joint objective and with their own objectives, meaning they are also in an approximate NE of the base game with either teacher. The full details of this proof are outlined in Appendix~\ref{appendix:theory}.

\section{Robustifying PLR}
In this section, we provide theoretical justification for the empirically observed effectiveness of PLR, and in the process, motivate a counterintuitive adjustment to the algorithm.

\label{sec:robust_plr}

\begin{figure}[h!] 
\vskip -0.1in
\begin{minipage}{\linewidth}
\begin{algorithm}[H]
{\small
\label{alg:plr+}
\SetAlgoLined
\caption{Robust PLR (PLR$^\bot$)}
Randomly initialize policy $\pi(\phi)$ and an empty level buffer, $\bm{\Lambda}$ of size $K$. \\
    \While{not converged}{
        Sample replay-decision Bernoulli, $d \sim P_{D}(d)$ \\
        \eIf{$d=0$}{
            Sample level $\theta$ from level generator\\
            Collect $\pi$'s trajectory $\tau$ on $\theta$, {\color{blue} with a stop-gradient $\phi_{\bot}$} \hfill {\color{gray} \emph{i.e. Suppress policy update}} 
        }
        {
        Use PLR to sample a replay level from the level store, $\theta \sim \bm{\Lambda}$ \\
        Collect policy trajectory $\tau$ on $\theta$ and update $\pi$ with rewards $\bm{R}(\tau)$
        }
        
        Compute PLR score, $S = \textbf{score}(\tau, \pi)$ \\
        Update $\bm{\Lambda}$ with $\theta$ using score $S$
    }
}
\end{algorithm}
\end{minipage}
\end{figure}

\subsection{Achieving Robustness Guarantees with PLR}
PLR provides strong empirical gains in generalization, but lacks any theoretical guarantees of robustness. One step towards achieving such a guarantee is to replace its L1 value-loss prioritizaton with a regret prioritization, using the methods we discuss in Section \ref{subsec:estimating_regret}: While L1 value loss may be good for quickly training the value function, it can bias the long-term training behavior toward high-variance policies. However, even with this change, PLR holds weaker theoretical guarantees because the random generating teacher can bias the student away from minimax regret policies and instead, toward policies that sacrifice robustness in order to excel in unstructured levels. We formalize this intuitive argument in the following corollary of Theorem \ref{theorem:main}.

\newtheorem{plr-corollary}{Corollary}
\begin{plr-corollary}
\label{corollary:plr}
Let $\overline{G}$ be the dual curriculum game in which the first teacher maximizes regret, so $U^1_t = U^R_t$, and the second teacher plays randomly, so  $U^2_t = U^U_t$.  Let $V^\theta(\pi)$ be bounded in $[B^-,B^+]$ for all $\theta, \pi$.  Further, suppose that $(\pi, \theta^1,\theta^2)$ is a Nash equilibrium of $\overline{G}$.  Let $R^* = \min_{\pi_A \in \Pi}\{\max_{\theta,\pi_B \in \Theta , \Pi}\{\textsc{Regret}^{\theta}(\pi_A,\pi_B)\}\}$ be the optimal worst-case regret.  Then $\pi$ is $2(B^+-B^-)(1-p)$ close to having optimal worst-case regret, or formally, $\max_{\theta,\pi_B \in \Theta , \Pi}\{\textsc{Regret}^{\theta}(\pi_A,\pi)\} \geq R^* - 2(B^+-B^-)(1-p)$.  Moreover, there exists environments for all values of $p$ within a constant factor of achieving this bound.
\end{plr-corollary}

The proof of Corollary \ref{corollary:plr} follows from a direct application of Theorem \ref{theorem:main} to show that a NE of $\overline{G}$ is an approximate NE for the base game of the first teacher, and through constructing a simple example where the student's best response in $\overline{G}$ fails to attain the minimax regret in $G$. These arguments are described in full in Appendix \ref{appendix:theory}.  This corollary provides some justification for why PLR improves robustness of the equilibrium policy, as it biases the resulting policy toward a minimax regret policy.  However, it also points a way towards further improving PLR: If the probability $p$ of using a teacher-generated level directly was set to $0$, then in equilibrium, the resulting policy converges to a minimax regret policy.
Consequently, we arrive at the counterintuitive idea of avoiding gradient updates from trajectories collected from randomly sampled levels, to ensure that at NE, we find a minimax regret policy. From a robustness standpoint, it is therefore optimal to train on less data. The modified PLR algorithm \plrabbrev{} with this counterintuitive adjustment is summarized in Algorithm \ref{alg:plr+}, in which this small change relative to the original algorithm is highlighted in {\color{blue} blue}. 

\subsection{Estimating Regret}
\label{subsec:estimating_regret}
In general, levels may differ in maximum achievable returns, making it impossible to know the true regret of a level without access to an oracle. As the L1 value loss typically employed by PLR does not generally correspond to regret, we turn to alternative scoring functions that better approximate regret. Two approaches, both effective in practice, are discussed below.

\textbf{Positive Value Loss\medspace}Averaging over all transitions with positive value loss amounts to estimating regret as the difference between maximum achieved return and predicted return on an episodic basis. However, this estimate is highly biased, as the value targets are tied to the agent's current, potentially suboptimal policy. As it only considers positive value losses, this scoring function leads to optimistic sampling of levels with respect to the current policy. When using GAE \citep{gae} to estimate bootstrapped value targets, this loss takes the following form, where $\lambda$ and $\gamma$ are the GAE and MDP discount factors respectively, and $\delta_t$, the  TD-error at timestep $t$:

\vspace{-0.2cm}
\[
\frac{1}{T}\sum_{t=0}^{T} \max \left(\sum_{k=t}^T(\gamma\lambda)^{k-t}\delta_k, 0\right).
\]

\textbf{Maximum Monte Carlo (MaxMC)\medspace} We can mitigate some of the bias of the positive value loss by replacing the value target with the highest return achieved on the given level so far during training. By using this maximal return, the regret estimates no longer depend on the agent's current policy. This estimator takes the simple form of $(1/T)\sum_{t=0}^{T} R_{\rm{max}} - V(s_t)$. In our dense-reward experiments, we compute this score as the difference between the maximum achieved return and $V(s_0)$. 
\michael{We should define RMAX}

\section{Replay-Enhanced PAIRED (\algoabbrev{})}
\label{sec:repaired}

We can replace the random generator teacher used by \plrabbrev{} with the PAIRED teacher. This extension entails a second student agent, the antagonist, also equipped with its own PLR level buffer.
In each episode, with probability $p$, the students evaluate their performances (but do not train) on a newly generated level and, with probability $1-p$, train on a level sampled from each student's own regret-prioritizing PLR buffer. Training only on the highest regret levels should mitigate inefficiencies in the PAIRED teacher's optimization procedure. We refer to this extension as \emph{Replay-Enhanced PAIRED} (REPAIRED). An overview of REPAIRED is provided by black arrows in Figure~\ref{figure:repaired_overview}, with the students being the protagonist and antagonist, while the full pseudocode is outlined in Appendix \ref{appendix:algorithms}.

Since \plrabbrev{} and PAIRED both promote regret in equilibrium, it would be reasonable to believe that the combination of the two does the same. A straightforward corollary of Theorem \ref{theorem:main}, which we describe in Appendix \ref{theorem:main}, shows that, in a theoretically ideal setting, combining these two algorithms as is done in \algoabbrev{} indeed finds minimax regret strategies in equilibrium.

\vspace{0.2cm}
\newtheorem{repaired-corollary}[]{Corollary}
{
\setcounter{repaired-corollary}{1}
\begin{repaired-corollary}
Let $\overline{G}$ be the dual curriculum game in which both teachers maximize regret, so $U^1_t = U^2_t = U^R_t$. Further, suppose that $(\pi, \theta^1,\theta^2)$ is a Nash equilibrium of $\overline{G}$. Then, $\pi \in \argmin_{\pi_A \in \Pi}\{\max_{\theta,\pi_B \in \Theta , \Pi}\{\textsc{Regret}^{\theta}(\pi_A,\pi_B)\}\}$.
\end{repaired-corollary}
}

This result gives us some amount of assurance that, if our method arrives at NE, then the protagonist has converged to a minimax regret strategy, which has the benefits outlined in~\citep{paired}: Since a minimax regret policy solves all solvable environments, whenever this is possible and sufficiently well-defined, we should expect policies resulting from the equilibrium behavior of \algoabbrev{} to be robust and versatile across all environments in the domain.

\section{Experiments}
\label{sec:experiments}

Our experiments firstly aim to (1) assess the empirical performance of the theoretically motivated \plrabbrev{}, and secondly, seek to better understand the effect of replay on unsupervised environment design, specifically (2) its impact on the zero-shot generalization performance of the induced student policies, and (3) the complexity of the levels designed by the teacher. To do so, we compare PLR and REPAIRED against their replay-free counterparts, DR and PAIRED, in the two highly distinct settings of discrete control with sparse rewards and continuous control with dense rewards. We provide environment descriptions alongside model and hyperparameter choices in Appendix \ref{appendix:exp_details}.

\subsection{Partially-Observable Navigation}
\label{sec:exp_minigrid}

Each navigation level is a partially-observable maze requiring student agents to take discrete actions to reach a goal and receive a sparse reward. Our agents use PPO~\citep{schulman2017proximal} with an LSTM-based recurrent policy to handle partial observability. Before each episode, the teacher designs the level in this order: beginning with an empty maze, it places one obstructing block per time step up to a predefined block budget, and finally places the agent followed by the goal.

\begin{figure}[h]
    \begin{minipage}{\textwidth}
    \centering\subfigure{\includegraphics[width=1\linewidth]{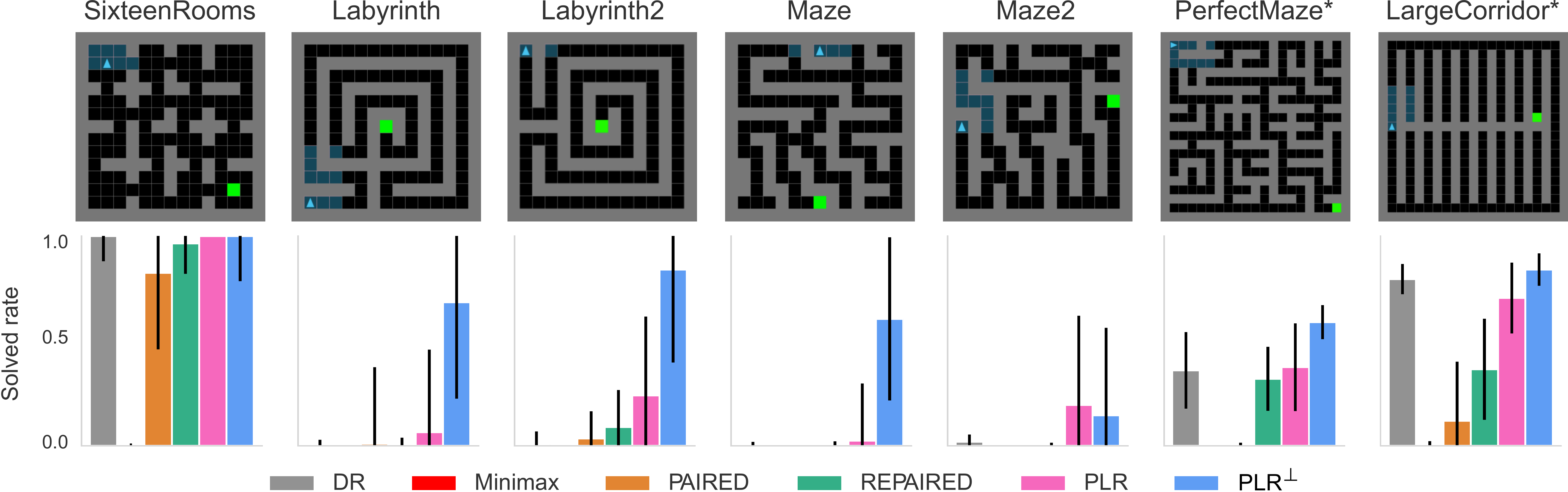}}  
    \caption{\small{Zero-shot transfer performance in challenging test environments after 250M training steps. The plots show median and interquartile range of solved rates over 10 runs. An asterisk (*) next to the maze name indicates the maze is procedurally-generated, and thus each attempt corresponds to a random configuration of the maze.}}
    \label{figure:main_minigrid}
    \end{minipage}
\end{figure}

\begin{figure}[t!]
    \begin{minipage}{\textwidth}
    \centering\subfigure{\includegraphics[width=1\linewidth]{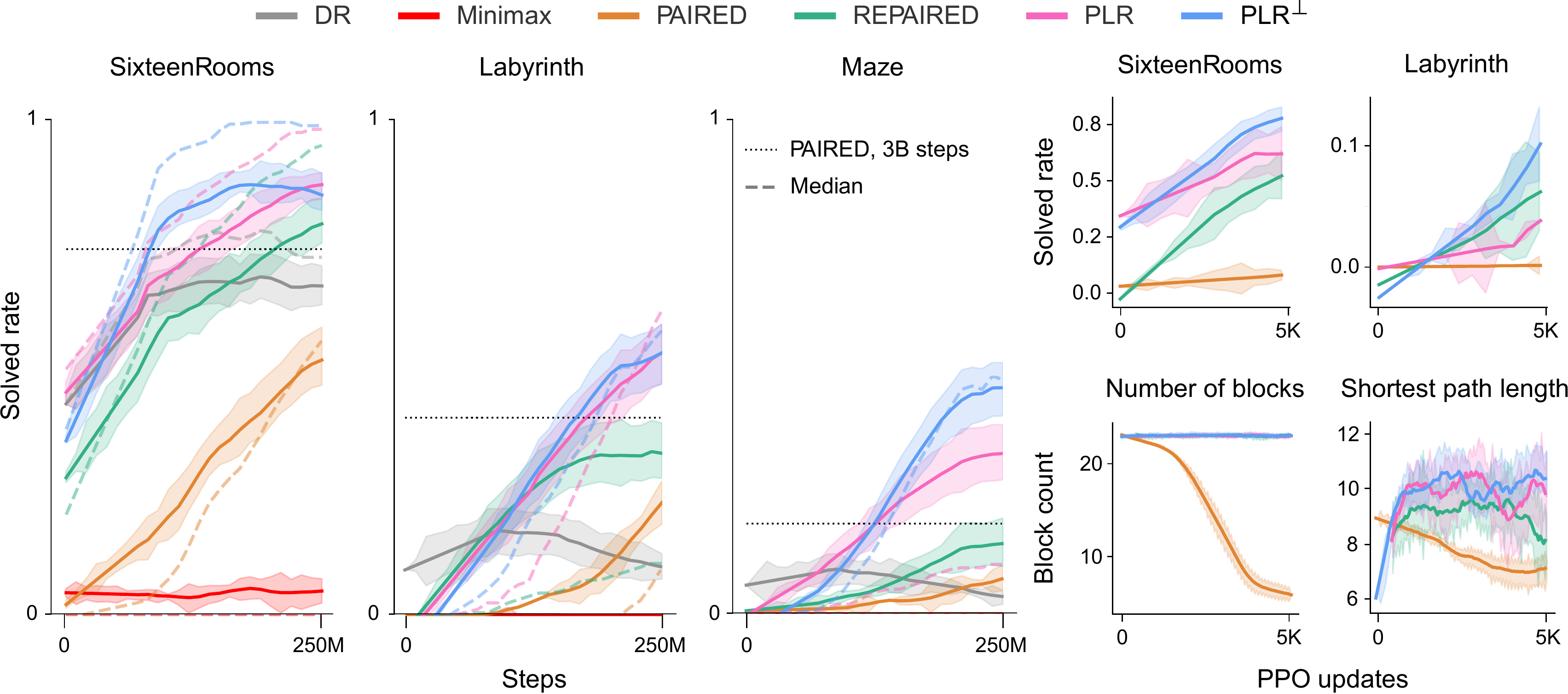}}  
    \caption{\small{Zero-shot transfer performance during training for PAIRED and REPAIRED variants. The plots show mean and standard error across 10 runs. The dotted lines mark the mean performance of PAIRED after 3B training steps, as reported in~\citep{paired}}, while dashed lines indicate median returns.}
    \label{figure:minigrid_curves}
    \end{minipage}
\end{figure}

\textbf{Zero-Shot Generalization}\medspace We train policies with each method for 250M steps and evaluate zero-shot generalization on several challenging OOD environments, in addition to levels from the full distribution of two procedurally-generated environments, PerfectMaze and LargeCorridor. We also compare against DR and minimax baselines. 
Our results in Figure~\ref{figure:main_minigrid} and~\ref{figure:minigrid_curves} show that \plrabbrev{} and REPAIRED both achieve greater sample-efficiency and zero-shot generalization than their replay-free counterparts. The improved test performance achieved by \plrabbrev over both DR and PLR when trained for an equivalent number of gradient updates, aggregated over all test mazes, is statistically significant ($p < 0.05$), as is the improved test performance of REPAIRED over PAIRED.  Well before 250 million steps, both PLR and \plrabbrev{} significantly outperform PAIRED after 3 billion training steps, as reported in~\citep{paired}. Further, both PLR variants lead to policies exhibiting greater zero-shot transfer than the PAIRED variants. Notably, the \plrabbrev{} agent learns to solve  mazes via an approximate right-hand rule. Table \ref{table:minigrid_eval} in Appendix \ref{appendix:subsec:minigrid} reports performance across all test mazes. The success of designing regret-maximizing levels via random search (curation) over learning a generator with RL suggests that for some UPOMDPs, the regret landscape, as a function of the free parameters $\theta$, has a low effective dimensionality \citep{bergstra2012random}. Foregoing gradient-based learning in favor of random search may then lead to faster adaptation to the changing regret landscape, as the policy evolves throughout training.

\begin{wrapfigure}{R}{0.4\textwidth}
    \vspace{-7mm} %
   \subfigure{\includegraphics[width=1\linewidth]{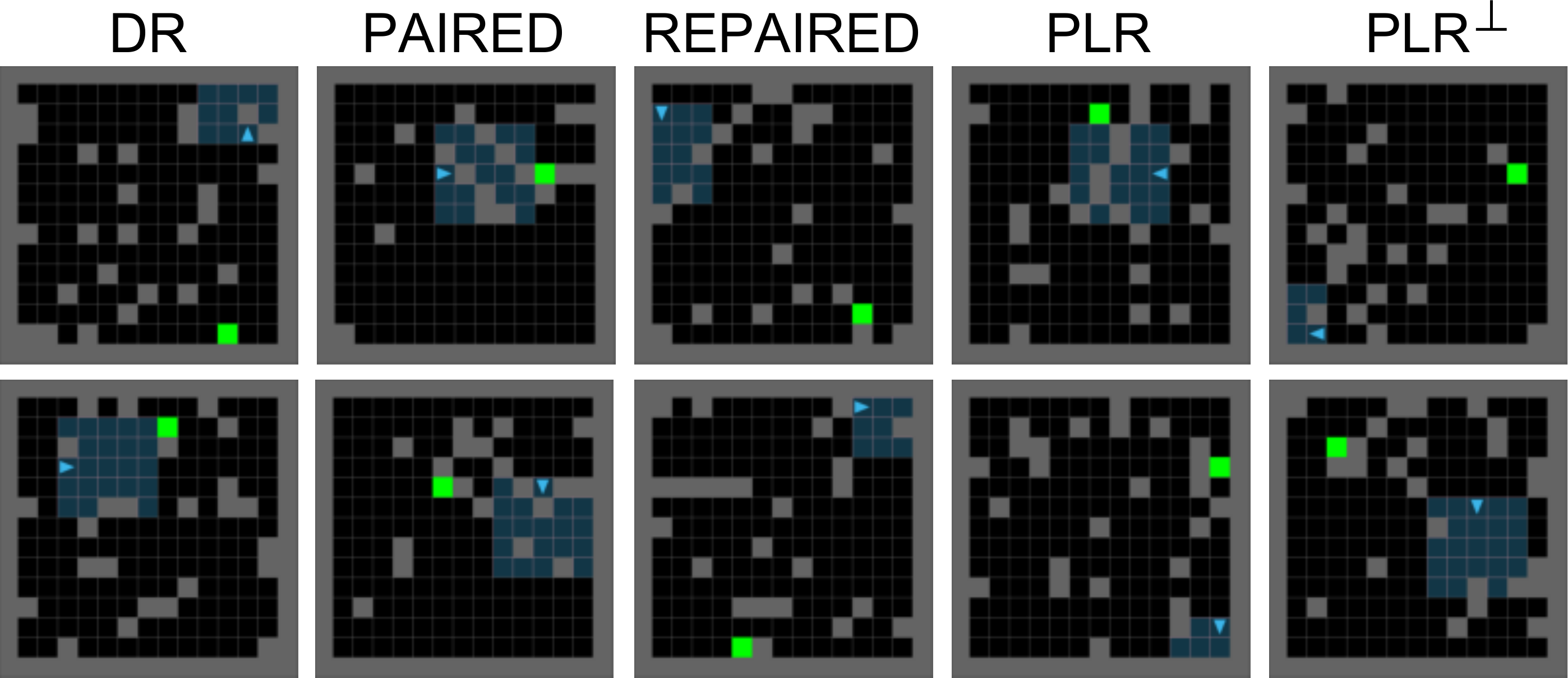}}
   \caption{Examples of emergent structures generated by each method.}
   \label{figure:minigrid_complexity_examples}
  \vspace{-.4cm}
\end{wrapfigure}

\textbf{Emergent Complexity}\medspace
As the student agents improve, the teachers must generate more challenging levels to maintain regret. We measure the resultant emergent complexity by tracking the number of blocks in each level and the shortest path length to the goal (where unsolvable levels are assigned a length of 0). These results, summarized in Figure~\ref{figure:minigrid_curves}, show that PAIRED slowly adapts the complexity over training while REPAIRED initially quickly grows complexity, before being overtaken by PAIRED. This more rapid onset of complexity may be due to REPAIRED's fast replay mechanism, and the long-term slowdown relative to PAIRED may be explained by its less frequent gradient updates. Our results over an extended training period in Appendix \ref{appendix:extra_exp} confirm that both PAIRED and REPAIRED slowly increase complexity over time, eventually matching that attained in just a fraction of the number of gradient steps by PLR and \plrabbrev{}. This result shows that random search is surprisingly efficient at continually discovering levels of increasing complexity, given an appropriate curation mechanism such as PLR. Figure \ref{figure:minigrid_complexity_examples} shows that, similar to methods with a regret-maximizing teacher, PLR finds levels exhibiting complex structure. 

\subsection{Pixel-Based Car Racing with Continuous Control}
\label{sec:exp_carracing}

\begin{figure}[h]
    \begin{minipage}{1\textwidth}
    \centering\subfigure{\includegraphics[width=1\linewidth]{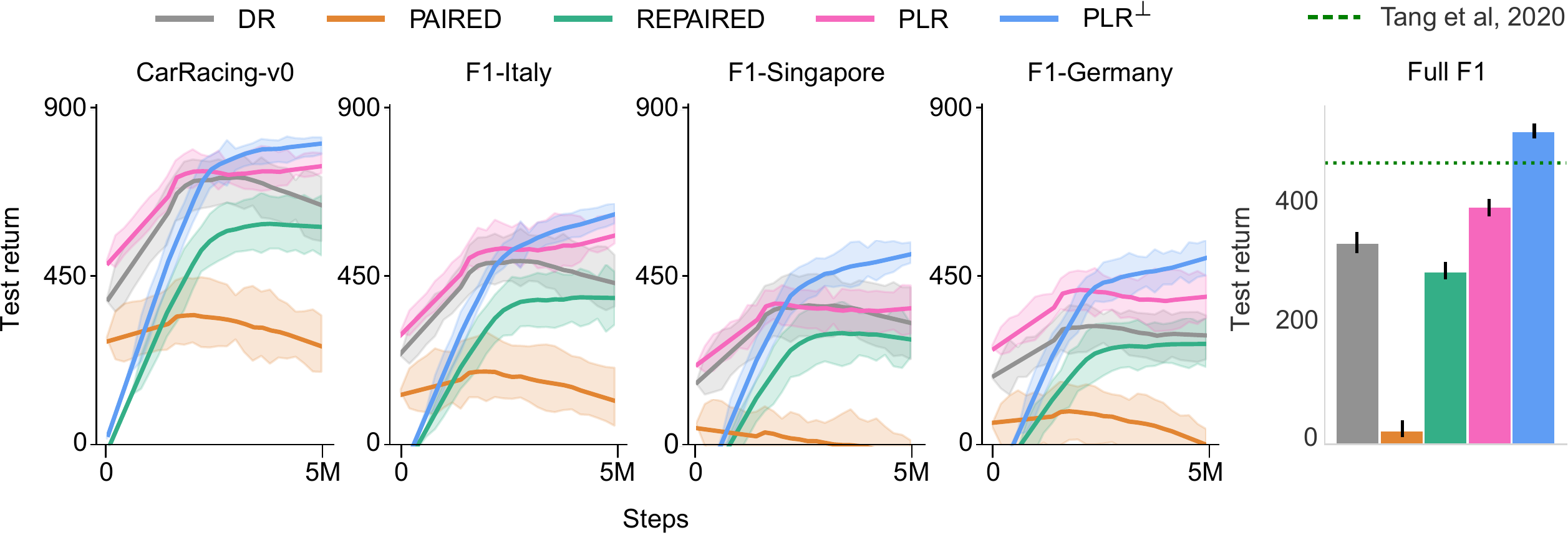}}
    \caption{\small{Zero-shot transfer performance. Plots show mean and standard error over 10 runs.}}
    \label{figure:carracing_main_results}
    \end{minipage}
\end{figure}

To test the versatility and scalability of our methods, we turn to an extended version of the CarRacing environment from OpenAI Gym~\citep{gym}. This environment entails continuous control with dense rewards, a 3-dimensional action space, and partial, pixel observations, with the goal of driving a full lap around a track. To enable UED of any closed-loop track, we reparameterize CarRacing to generate tracks as B{\'e}zier curves~\citep{bezier_ref} with arbitrary control points. The teacher generates levels by choosing a sequence of up to 12 control points, which uniquely defines a B{\'e}zier track within specific, predefined curvature constraints. After 5M steps of training, we test the zero-shot transfer performance of policies trained by each method on 20 levels replicating official human-designed Formula One (F1) tracks (see~\Cref{figure:f1_tracks} in the Appendix for a visualization of the tracks). Note that these tracks are significantly OOD, as they cannot be defined with just 12 control points. In Figure~\ref{figure:carracing_main_results} we show the progression of zero-shot transfer performance for the original CarRacing environment, as well as three F1 tracks of varying difficulty, while also including the final performance on the full F1 benchmark. For the final performance, we also evaluated the state-of-the-art CarRacing agent from \citep{attentionagent} on our new F1 benchmark. 

Unlike in the sparse, discrete navigation setting, we find DR leads to moderately successful policies for zero-shot transfer in CarRacing. Dense rewards simplify the learning problem and random Bezier tracks occasionally contain the challenges seen in F1 tracks, such as hairpin turns and observations showing parallel tracks due to high local curvature. Still, we see that policies trained by selectively sampling tracks to maximize regret significantly outperform those trained by uniformly sampling from randomly generated tracks, in terms of zero-shot transfer to the OOD F1 tracks. Remarkably, with a replay rate of 0.5, \plrabbrev{} sees statistically significant ($p < 0.001$) gains over PLR in zero-shot performance over the full F1 benchmark, despite directly training on only half the rollout data using half as many gradient updates. Once again, we see that random search with curation via PLR produces a rich  selection of levels and an effective curriculum.  

We also observe that PAIRED struggles to train a robust protagonist in CarRacing. Specifically, PAIRED overexploits the relative strengths of the antagonist over the protagonist, finding curricula that steer the protagonist towards policies that ultimately perform poorly even on simple tracks, leading to a gradual reduction in level complexity. We present training curves revealing this dynamic in Appendix \ref{appendix:extra_exp}. As shown in Figure~\ref{figure:carracing_main_results}, REPAIRED mitigates this degeneracy substantially, though not completely, inducing a policy that significantly outperforms PAIRED ($p < 0.001$) in mean performance on the full F1 benchmark, but underperforms DR. Notably, \plrabbrev{} exceeds the performance of the state-of-the-art AttentionAgent~\citep{attentionagent}, despite not using a self-attention policy and training on less than 0.25\% of the number of environment steps in comparison. These gains come purely from the induced curriculum.  Figure~\ref{figure:carracing_min_returns} in Appendix \ref{appendix:extra_exp} further reveals that \plrabbrev{} produces CarRacing policies that tend to achieve higher minimum returns on average compared to the baseline methods, providing further evidence of the benefits of the minimax regret property coupled with a fast replay-based mechanism for efficiently finding high-regret levels.

\section{Related Work}
In inducing parallel curricula, DCD follows a rich lineage of curriculum learning methods~\citep{bengio_curriculum, schmidhuber_curriculum, curriculum_rl_survey2, curriculum_rl_survey1}. Many previous curriculum learning algorithms resemble the curator in DCD, sharing similar underlying selective-sampling mechanisms as \plrabbrev{}. Most similar is TSCL~\citep{tscl}, which prioritizes levels based on return rather than value loss, and has been shown to overfit to training levels in some settings \citep{plr}. In our setting, replayed levels can be viewed as past strategies from a level-generating teacher. This links our replay-based methods to fictitious self-play~\cite[FSP,][]{fictitious_sp}, and more closely, Prioritized FSP \citep{vinyals2019grandmaster}, which selectively samples opponents based on historic win ratios.

Recent approaches that make use of a generating adversary include Asymmetric Self-Play~\citep{sukhbaatar2018intrinsic, openai2021asymmetric}, wherein one agent proposes tasks for another in the form of environment trajectories, and AMIGo~\citep{amigo}, wherein the teacher is rewarded for proposing reachable goals. While our methods do not presuppose a goal-based setting, others have made progress here using generative modeling~\citep{goalgan, Racaniere2020Automated}, latent skill learning~\citep{carml}, and exploiting model disagreement~\citep{NEURIPS2020_566f0ea4}. These methods are less generally applicable than \plrabbrev{}, and unlike our DCD methods, they do not provide well-principled robustness guarantees.

Other recent algorithms can be understood as forms of UED and like DCD, framed in the lens of decision theory. POET~\citep{poet, enhanced_poet}, a coevolutionary approach~\citep{Popovici2012}, uses a population of \emph{minimax} (rather than minimax regret) adversaries to construct terrain for a BipedalWalker agent. In contrast to our methods, POET requires training a large population of both agents and environments and consequently, a sizable compute overhead. APT-Gen~\citep{fang2021adaptive} also procedurally generates tasks, but requires access to target tasks, whereas our methods seek to improve zero-shot transfer.

The DCD framework also encompasses adaptive domain randomization methods~\cite[DR,][]{mehta2019activedomain,dr_evolutionary}, which have seen success in assisting sim2real transfer for robotics~\citep{domain_randomization, james2017transferring, dexterity, rubics_cube}. DR itself is subsumed by procedural content generation \citep{risi_togelius_pcg}, for which UED and DCD may be seen as providing a formal, decision-theoretic framework, enabling development of provably optimal algorithms.

\section{Discussion}
\label{sec:conclusion}
We established a novel connection between PLR and minimax regret UED approaches like PAIRED, by developing the theory of Dual Curriculum Design (DCD). In this setting, a student policy is challenged by a team of two co-adapting, regret-maximizing teachers: one, a generator that creates new levels, and the other, a curator that selectively samples previously generated levels for replay. This view unifies PLR and PAIRED, which are both instances of DCD. Our theoretical results on DCD then enabled us to prove that PLR attains a minimax regret policy at NE, thereby providing the first theoretical characterization of the robustness of PLR. Notably our theory leads to the counterintuitive result that PLR can be made provably robust by training on less data, specifically, by only using the trajectories on levels sampled for replay. In addition, we developed Replay-Enhanced PAIRED (REPAIRED), which extends the selective replay-based updates of \plrabbrev{} to PAIRED, and proved it shares the same robustness guarantee at NE.
Empirically, in two highly distinct environments, we found that \plrabbrev{} significantly improves zero-shot generalization over PLR, and REPAIRED, over PAIRED. As our methods solely modify the order of levels visited during training, they can, in principle, be combined with many other RL methods to yield potentially orthogonal improvements in sample-efficiency and generalization.

While these DCD-based improvements to PLR and PAIRED empirically lead to more robust policies, it is important to emphasize that our theoretical results only prove a minimax regret guarantee at NE for these methods; however, they provide no explicit guarantee of convergence to such NE. Further, it is worth highlighting that replay-based methods like \plrabbrev{} are completely dependent on the quality of levels proposed by the generator. Our results show that simply curating high regret levels discovered via random search is enough to outperform the RL-based PAIRED teacher in the domains studied. We expect that advancing methods for defining or adapting the generator's proposal distribution holds great potential to improve the efficacy of our methods, especially in more complex, higher-dimensional domains, where random search may prove ineffective for finding useful training levels. Importantly, our methods assume an appropriate choice of what constitutes the UPOMDP's free parameters. Our methods cannot be expected to produce robust policies for zero-shot transfer if the set of environments defined by the free parameters does not sufficiently align with the transfer domain of interest. Designing the environment parameterization for successful zero-shot transfer to a specific target domain can be highly non-trivial, posing an important problem for future research.

Looking beyond environment design, we notice that long-running UED processes in expansive UPDOMPs closely resemble continual learning in open-ended domains. The congruency of these settings suggests our contributions around DCD may extend to more general continual learning problems in which agents must learn to master a diverse sequence of tasks with predefined (or inferred) episode boundaries---if tasks are assumed to be designed by a regret-maximizing teacher. Thus, DCD-based methods like \plrabbrev{} may yield more general policies for continual learning. We anticipate many exciting crossovers between these areas of research in the years to come.

\impactsection{
Given the rapid progress in applying RL to ever more complex domains, we can confidently expect a continued rise in real-world deployments of RL systems in the coming years. Unlike in simulation, in the real world, the environment tends to exhibit much more variability, which may not be explicitly coded into the associated simulator used for training. Deployed RL agents are thus liable to make many mistakes due to unexpected environment variations. Our methods for improving UED lead to more robust RL agents across a potentially wide range of changes to the environment. Thus, our work may prove to be a useful tool in attaining safer, more reliable RL agents, helping to enable the application of RL to more real-world problems. 

By increasing the applicability of RL to real-world settings, our work may exacerbate the more general risks of deploying machine learning: increased unemployment; overreliance on biased models that potentially reinforce common misconceptions and societal inequalities; and the advancement of automated weapons. Particular to our methods, as discussed in Section \ref{sec:conclusion}, aligning the choice of free parameters for the UPOMDP to the target domain of interest is important for successful transfer. While this choice of free parameters then acts as a potential point of failure, the UPOMDP abstraction underlying UED reveals this problem generally impacts all RL methods aiming to train robust policies; a UPOMDP simply makes explicit the otherwise implicit space of environment configurations defined by a standard POMDP. By forcing us to consider where our training environment departs from reality, UED methods encourage designing RL systems in a way that is more aware of the underlying assumptions about the environment, thereby leading to more principled, robust systems.
}

\acksection{
We would like to thank Natasha Jaques, Patrick Labatut, and Heinrich K\"{u}ttler for fruitful discussions that helped inform this work. Further, we are grateful to our anonymous reviewers for their valuable feedback. MJ is supported by the FAIR PhD program. This work was funded by Facebook.
}

\clearpage

\bibliographystyle{abbrv}
\bibliography{refs.bib}

\newpage

\appendix

\section{Theoretical Results}
\label{appendix:theory}
In this section we prove the theoretical results around the dual curriculum game and use these results to show approximation bounds for our methods, given that they have reached a Nash equilibrium (NE).

The first theorem is the main result that allows us to analyze dual curriculum games. The high-level result says that the NE of a dual curriculum game are approximate NE of the base game from the perspective of any of the individual players, or from the perspective of the joint strategy.

\setcounter{main_theorem}{0}
\begin{main_theorem}
Let $B$ be the maximum difference between $U^1_t$ and $U^2_t$, and let $(\pi, \theta^1, \theta^2)$ be a NE for $\overline{G}$.  Then $(\pi, p\theta^1+(1-p)\theta^2)$ is an approximate NE for the base game with either teacher or for a teacher optimizing their joint objective. More precisely, it is a $2Bp(1-p)$-approximate NE when $U_t = pU^1_t+(1-p)U^2_t$, a $2B(1-p)$-approximate NE when $U_t = U^1_t$, and a $2Bp$-approximate NE when $U_t = U^2_t$.
\end{main_theorem}

At a high level, this is true because, for low values of $p$, the best-response strategies for the individual players can be thought of as approximate-best response strategies for the joint-player, and vis-versa.  Since the Nash Equilibrium consists of each of the players playing their own best response, they must be playing an approximate best response for the joint-player. We provide a formal proof below:

\begin{proof}
Let $B$ be the maximum difference between $U^1_t$ and $U^2_t$, and let $(\pi, \theta^1, \theta^2)$ be a Nash Equilibrium for $\overline{G}$.  Then consider $p\theta^1+(1-p)\theta^2$ as a strategy in the base game for the joint player $pU^1_t+(1-p)U^2_t$.  Let $\theta^{1+2}$ be the best response for the joint player to $\pi$. Since $\pi$ is a best response by assumption, it is sufficient to show that $p\theta^1+(1-p)\theta^2$ is an approximate best response. We then have

\begin{align}
    &U_t(\pi,p\theta^1+(1-p)\theta^2)
    \\
    =& p^2U^1_t(\pi,\theta^1)+p(1-p)U^2_t(\pi,\theta^1)+p(1-p)U^1_t(\pi,\theta^2)+(1-p)^2U^2_t(\pi,\theta^2)\\
    \geq& p^2U^1_t(\pi,\theta^1)+p(1-p)(U^1_t(\pi,\theta^1)-B)+p(1-p)(U^2_t(\pi,\theta^2)-B)+(1-p)^2U^2_t(\pi,\theta^2)\\
    =& pU^1_t(\pi,\theta^1)+(1-p)U^2_t(\pi,\theta^2)-2Bp(1-p)\\
    \geq& U_t(\pi,\theta^{1+2})-2Bp(1-p)
\end{align}

Thus, we have shown that $(\pi,p\theta^1+(1-p)\theta^2)$ represents an $2Bp(1-p)$-Nash equilibrium for the joint player. For the first teacher we have the opposite condition trivially, the teacher is doing a best response to the student. We must now show that the student is doing an approximate best response to the teacher.

Let $\pi^{1}$ be the best response to the first teacher (with utility $U^1_t$) and let $\pi^{1+2}$ be the best response policy to the joint teacher.  In this argument we will start with the observation that $U_s(\pi^1, \theta^{1+2}) \leq U_s(\pi^{1+2}, \theta^{1+2})$ by definition, and then argue that we can construct an upper bound on the performance of $\pi^1$ on $\theta^1$, $U_s(\pi^1,\theta^1)$, and a lower bound on the performance of $\pi^{1+2}$ on $\theta^1$, $U_s(\pi^{1+2},\theta^1)$.  We get the desired result by combining these two arguments.

First we use $U_s(\pi^1, \theta^{1+2})$ to upper bound $U_s(\pi^1,\theta^1)$: 

\begin{align}
    U_s(\pi^1, \theta^{1+2}) &= p U_s(\pi^1, \theta^1) + (1-p) U_s(\pi^1, \theta^2) \\
    &\geq p U_s(\pi^1, \theta^1) + (1-p) (U_s(\pi^1, \theta^1) - B) \\
    &= U_s(\pi^1, \theta^1) - (1-p) B
\end{align}

Second we can use $U_s(\pi^{1+2}, \theta^{1+2})$ to lower bound $U_s(\pi^{1+2},\theta^1)$:

\begin{align}
    U_s(\pi^{1+2}, \theta^{1+2}) &= p U_s(\pi^{1+2}, \theta^1) + (1-p) U_s(\pi^{1+2}, \theta^2) \\
    &\leq p U_s(\pi^{1+2}, \theta^1) + (1-p) (U_s(\pi^{1+2}, \theta^1) + B) \\
    &= U_s(\pi^{1+2}, \theta^1) + (1-p) B
\end{align}

Putting this all together, we have
\[
 U_s(\pi^{1+2}, \theta^1) + (1-p) B \geq U_s(\pi^1, \theta^1) - (1-p) B. 
\]

Which, after rearranging terms, gives 
\[
 U_s(\pi^{1+2}, \theta^1) \geq U_s(\pi^1, \theta^1) - 2 (1-p) B
\]
as desired.  Repeating the symmetric argument shows the desired property for the second teacher.
\end{proof}

Following this main theorem, we can apply it to two of our methods. First we can apply it to naive PLR, which trains on a mixture of domain randomization (a teacher with utility $U_t^C$) and the PLR bandit (a teacher with utility $U_t^R$). This result shows that as we reduce the number of random episodes, the approximation to a minimax regret strategy improves. The intuition behind this is a direct application of Theorem \ref{theorem:main}, to show that it is an approximate Nash for the minimax regret player, and then showing that the minimax reget player has access to a strategy which ensures small regret, thus the regret that the equilibrium ensures must be approximately small.

\setcounter{plr-corollary}{0}
\begin{plr-corollary}
Let $\overline{G}$ be the dual curriculum game in which the first teacher maximizes regret, so $U^1_t = U^R_t$, and the second teacher plays randomly, so  $U^2_t = U^U_t$.  Let $V^\theta(\pi)$ be bounded in $[B^-,B^+]$ for all $\theta, \pi$.  Further, suppose that $(\pi, \theta^1,\theta^2)$ is a Nash equilibrium of $\overline{G}$.  Let $R^* = \min_{\pi_A \in \Pi}\{\max_{\theta,\pi_B \in \Theta , \Pi}\{\textsc{Regret}^{\theta}(\pi_A,\pi_B)\}\}$ be the optimal worst-case regret.  Then $\pi$ is $2(B^+-B^-)(1-p)$ close to having optimal worst-case regret, or formally, $\max_{\theta,\pi_B \in \Theta , \Pi}\{\textsc{Regret}^{\theta}(\pi_A,\pi)\} \geq R^* - 2(B^+-B^-)(1-p)$.  Moreover, there exists environments for all values of $p$ within a constant factor of achieving this bound.
\end{plr-corollary}

\begin{proof}
Since $V^\theta(\pi)$ is bounded in $[B^-,B^+]$ for all $\theta, \pi$, we know that  $U^1_t$ and $U^2_t$ are within $(B^+ - B^-)$ of each other.  Thus by Theorem \ref{theorem:main} we have that $(\pi, \theta^1,\theta^2)$ is a $2(B^+ - B^-)(1-p)$-Nash equilibrium of the base game when $U_t = U_t^1$. Thus $\pi$ is a $2(B^+ - B^-)(1-p)$ approximate best-response to $\theta^1$.  However, since $\theta^1$ is a best response it chooses a regret maximizing parameter distribution.  Thus the $2(B^+ - B^-)(1-p)$ does not just measure the sub-optimally of $\pi$ with respect to $\theta^1$, but measures the worst-case regret of $\pi$ across all $\theta$ as desired.

The intuition for the existence of examples in which this approximation of regret decays linearly in $p$ is that a random level and the maximal regret level can be very different, and so the two measures may diverge drastically.  For an example environment where $\pi$ deviates strongly from the minimax regret strategy, consider the one-step UMDP described in Table \ref{table:edge_case_game}.

\begin{table}[h!]
\begin{center}
\begin{tabular}{ c | c | c  | c  }
 & $\theta_0$ & $\theta_1$ & $\theta_2 \dotsc \theta_n$ \\
\hline
$\pi_0$ & $B$ & $0$ & $0$ \\
\hline
$\pi_1$ & $0$ & $B$ & $0$ \\
\hline
$\pi_2$ & $Bp+2\epsilon$ & $0$ & $\frac{Bp}{2}+\epsilon$ \\
\hline
$\pi_3$ & $0$ & $Bp+2\epsilon$ & $\frac{Bp}{2}+\epsilon$ \\
\end{tabular}
\caption{In this environment all payoffs are between $0$ and $B$(for $p \in (0,1)$ and $\epsilon<\frac{B(1-p)}{2}$), where $B$ is assumed to be positive.  Randomizing between $\pi_0$ and $\pi_1$ minimizes regret, but choosing $\pi_2$ or $\pi_3$ is better in expectation under the uniform distribution.  For large $n$ it is especially clear that $\pi_2$ and $\pi_3$ have better expected value under the uniform distribution, though we show that even for $n=2$, the optimal joint policy can mix between $\pi_2$ and $\pi_3$ incurring high regret.}
\label{table:edge_case_game}
\end{center}
\end{table}

Note that in Table \ref{table:edge_case_game}, no policy has less than $\frac{B}{2}$ regret, since every policy will have to incur $B$ regret on either $\{\theta_0,\theta_1\}$ at least half the time. The minimax regret policy mixes uniformly between $\pi_0$ and $\pi_1$ to achieve regret of exactly $\frac{B}{2}$.  We can ignore $\theta_2 \dotsc \theta_n$ for the regret calculations by assuming that $\epsilon < \frac{B(1-p)}{2}$, since every policy achieves less than $\frac{B}{2}$ regret on these levels.

Our claim is that in equilibrium of $\overline{G}$ in this environment, the student policy can incur $\frac{B}{2} + \frac{B(1-p)}{2}-\epsilon$ regret, $\frac{B(1-p)}{2}- \epsilon$ more than the minimax regret policy.  An example of such an equilibrium point would be when the student policy uniformly randomizes between $\pi_2$ and $\pi_3$, which we will call $\pi_{2+3}$,  when the minimax teacher uniformly randomizes between $\theta_{0}$ and $\theta_{1}$ which we will call $\theta_{0+1}$, and when the uniform teacher randomizes exactly which we call $\tilde{\theta}$.  To check this we must show that $(\pi_{2+3}, \theta_{0+1}, \tilde{\theta})$ is in fact a NE of $\overline{G}$.  Then we must show that $\pi_{2+3}$ incurs $\frac{B}{2} + \frac{B(1-p)}{2}-\epsilon$ regret.  

To show that $(\pi_{2+3}, \theta_{0+1}, \tilde{\theta})$ is a NE of $\overline{G}$ first note that $\tilde{\theta}$ is trivially a best response for the uniform utility function.  Also note that $\theta_{0+1}$ maximizes the regret of $\pi_{2+3}$ since $\theta_0$ and $\theta_1$ are the only two parameters on which $\pi_{2+3}$ incur regret, and they incur the same regret; thus, any mixture over them will be optimal for the regret-based teacher.  Finally, we need to show that $\pi_{2+3}$ is optimal for the student.  To do this we will calculate the expected value of each policy and notice that the expected values for $\pi_2$ and $\pi_3$ are higher than for $\pi_0$ and $\pi_1$.  Thus any optimal policy will place no weight on $\pi_0$ and $\pi_1$, but any distribution over $\pi_2$ and $\pi_3$ will be equivalently optimal. 
By symmetry, we can show only the calculations for $\pi_0$ and $\pi_2$:
\begin{align}
    \pi_0 &= p(\frac{1}{2}B + \frac{1}{2}0)+ (1-p) 0 = \frac{Bp}{2} \\
    \pi_2 &= p(\frac{1}{2} (Bp+2\epsilon) + \frac{1}{2}0) + (1-p)(\frac{Bp}{2}+\epsilon) = \frac{Bp}{2}+\epsilon
\end{align}

Thus $\pi_2$ and $\pi_3$ achieve $\epsilon$ higher expected value by the joint distribution.  Thus, we know that $\pi_{2+3}$ is a best response and $(\pi_{2+3}, \theta_{0+1}, \tilde{\theta})$ is in fact a NE of $\overline{G}$.   

Finally, we simply need to show that $\pi_{2+3}$ incurs $\frac{B}{2} + \frac{B(1-p)}{2}-\epsilon$ regret. WLOG, we can evaluate its regret on $\theta_0$.  On $\theta_0$, $\pi_{2+3}$ achieves $\frac{Bp}{2}+\epsilon$ reward while $\pi_0$ achieves $B$.  Thus $\pi_{2+3}$ incurs regret of $B - (\frac{Bp}{2}+\epsilon) = \frac{B}{2} + \frac{B - Bp}{2}-\epsilon = \frac{B}{2} + \frac{B(1-p)}{2}-\epsilon$ as desired.  As discussed before, since the minimax regret policy achieves $\frac{B}{2}$, this is $\frac{B(1-p)}{2}-\epsilon$ more regret than optimal. 
\end{proof}

Lastly, we can also apply Theorem \ref{theorem:main} to prove that REPAIRED achieves a minimax regret strategy in equilibrium.  The intuition behind this corollary is that, since the utility functions of both teachers are the same, the approximate NE ensured by Theorem \ref{theorem:main} is actually a true NE; therefore, the minimax theorem applies.  
\setcounter{repaired-corollary}{1}
\begin{repaired-corollary}
Let $\overline{G}$ be the dual curriculum game in which both teachers maximize regret, so $U^1_t = U^2_t = U^R_t$. Further, suppose that $(\pi, \theta^1,\theta^2)$ is a Nash equilibrium of $\overline{G}$. Then, $\pi \in \argmin_{\pi_A \in \Pi}\{\max_{\theta,\pi_B \in \Theta , \Pi}\{\textsc{Regret}^{\theta}(\pi_A,\pi_B)\}\}$.
\end{repaired-corollary}
\begin{proof}
Since $U^1_t = U^2_t = U^R_t$ the joint objective is $p U^1_t + (1-p)U^2_t = U^R_t$.  Note that since $U^1_t = U^2_t$, $B=0$.  Thus by Theorem \ref{theorem:main} $(\pi, p\theta^1+(1-p)\theta^2)$ is a $0$-Nash Equilibrium of the base game with teacher objective $U^R_t$, thus by the minimax theorem, $\pi \in \argmin_{\pi_A \in \Pi}\{\max_{\theta,\pi_B \in \Theta , \Pi}\{\textsc{Regret}^{\theta}(\pi_A,\pi_B)\}\}$ as desired.
\end{proof}

\clearpage

\section{Algorithms}
\label{appendix:algorithms}

Although the PLR update rule for the level buffer of size $K$ in the case of unbounded training levels is described in \citep{plr}, we provide the pseudocode for this update rule in Algorithm \ref{alg:plr_update_rule} for completeness. Given staleness coefficient $\rho$, temperature $\beta$, a prioritization function $h$ (e.g. rank), level buffer scores $S$, level buffer timestamps $C$, and the current episode count $c$ (i.e. current timestamp), the $P_{\textnormal{replay}}$ update takes the form
\begin{align*}
    P_{\text{replay}} &= (1-\rho) \cdot P_S + \rho \cdot P_C, \\
    P_S &= \frac{h(S_i)^{1/\beta}}{\sum_j h(S_j)^{1/\beta}}, \\
    P_C &= \frac{c - C_i}{\sum_{C_j \in C} c - C_j}.
\end{align*}

The pseudocode for Replay-Enhanced PAIRED (REPAIRED), the method described in Section~\ref{sec:repaired}, is presented in Algorithm~\ref{alg:repaired}.  

\begin{figure}[h!] 
\vskip -0.1in
\begin{minipage}{\linewidth}
\begin{algorithm}[H]
{
\label{alg:plr_update_rule}
\SetAlgoLined
\caption{PLR level-buffer update rule}
    \KwIn{Level buffer $\bm{\Lambda}$ of size $K$ with scores $S$ and timestamps $C$; level $\theta$; level score $S_{\theta}$; and current episode count $c$}
    \eIf{$|\bm{\Lambda}| < K$}{
        Insert $\theta$ into $\bm{\Lambda}$, and set $S(\theta) = S_{\theta}$, $C(\theta) = c$ \\
    }
    {
        Find level with minimal support,    $\theta_{\textnormal{min}} = \underset{\theta}{\arg\min\;}  P_{\textnormal{replay}}(\theta)$ \\
        \If{$S(\theta_{\textnormal{min}}) < S_{\theta}$}{
            Remove $\theta_{\textnormal{min}}$ from $\bm{\Lambda}$ \\
            Insert $\theta$ into $\bm{\Lambda}$, and set $S(\theta) = S_{\theta}$, $C(\theta) = c$ \\
            Update $P_{\textnormal{replay}}$ with latest scores $S$ and timestamps $C$ \\
        }
    }
}
\end{algorithm}
\end{minipage}
\end{figure}

\begin{figure}[h!] 
\vskip -0.1in
\begin{minipage}{\linewidth}
\begin{algorithm}[H]
{
\label{alg:repaired}
\SetAlgoLined
\caption{REPAIRED}

Randomly initialize Protagonist, Antagonist, and Generator policies $\pi^A(\phi^A)$, $\pi^B(\phi^B)$, and $\tilde{\theta}$ \\
Initialize Protagonist and Antagonist PLR level buffers $\bm{\Lambda}^A$ and $\bm{\Lambda}^B$ \\
    \While{not converged}{
        Sample replay-decision Bernoulli, $d \sim P_{D}(d)$ \\
        \eIf{$d = 0$}{
            Teacher policy $\tilde{\theta}$ generates the next level, $\theta$ \\
            Set $\theta^A = \theta^B = \theta$ \\
            Collect trajectory $\tau^A$ on $\theta^A$ and $\tau^B$ on $\theta^B$ with stop-gradients $\phi^A_{\bot}$, $\phi^B_{\bot}$ \\
            Update $\tilde{\theta}$ with $\textsc{Regret}^{\theta}(\pi^A, \pi^B)$ \\
        } 
        {
            PLR samples replay levels, $\theta^A \sim \bm{\Lambda}^A$ and $\theta^B \sim \bm{\Lambda}^B$ \\
            Collect trajectory $\tau^A$ on $\theta^A$ and $\tau^B$ on $\theta^B$ \\
            Update $\pi^A$ with rewards $\bm{R}(\tau^A)$, and $\pi^B$, with rewards $\bm{R}(\tau^B)$ \\
        }

        Compute PLR score $S^A = \textbf{score}(\tau^A, \tau^B, \pi^A)$ \\ Compute PLR score $S^B = \textbf{score}(\tau^B, \tau^A, \pi^B)$ \\
        Update $\bm{\Lambda}^A$ with $\theta^A$ using score $S^A$ \\
        Update $\bm{\Lambda}^B$ with $\theta^B$ using score $S^B$ \\
    }
}
\end{algorithm}
\end{minipage}
\end{figure}

\clearpage

\section{Additional Experimental Results}
\label{appendix:extra_exp}
This section provides additional experimental results in MiniGrid and CarRacing environments. Note that we determine the statistical significance of our results using a Welch t-test \citep{welch1947generalization}.

\subsection{Extended Results for MiniGrid}
\label{appendix:subsec:minigrid}

Unlike the original maze experiments used to evaluate PAIRED in \citep{paired}, we conduct our main maze experiments with a block budget of 25 blocks (reported in Section \ref{sec:exp_minigrid}), rather than 50 blocks. Following the environment parameterization in \citep{paired}, for a block budget of $B$, the teacher attempts to place $B$ blocks that act as obstacles when designing each maze level. However, the teacher can place fewer than $B$ blocks, as placing a block in a location already occupied by a block results in a no-opt. We found that PAIRED underperforms DR when both methods are given a budget of 50 blocks, a setting in which randomly sampled mazes exhibit enough structural complexity to allow DR to learn highly robust policies. Note that \citep{paired} used a DR baseline with a 25-block budget. With a 50-block budget, DR and all replay-based methods are able to fully solve almost all test mazes after around 500M steps of training, making UED of mazes with a 50-block budget too simple of a setting to provide an informative comparison among the methods studied.

\subsubsection{Mazes with a 25-block budget}
\label{appendix:subsubsec:minigrid_25b}

\begin{figure}[t!]
    \begin{minipage}{\textwidth}
    \subfigure{\includegraphics[width=\linewidth]{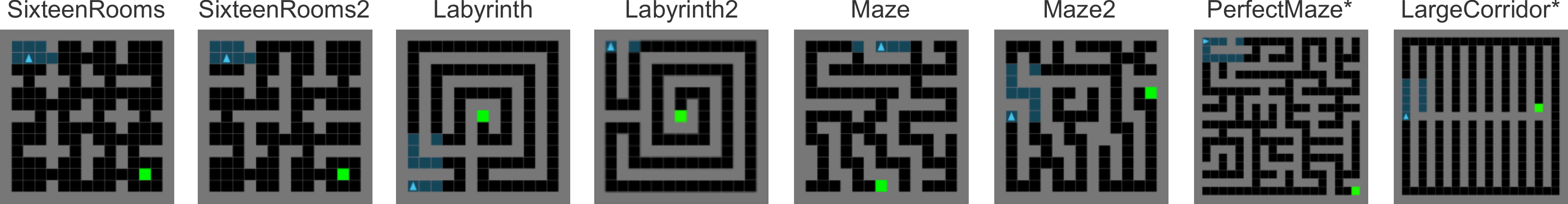}}
    \caption{\small{Test maze environments for evaluating zero-shot transfer. An asterisk (*) next to the maze name indicates the maze is procedurally-generated, and thus each attempt corresponds to a random configuration of the maze.}}
    \label{figure:mg_test_mazes}
    \end{minipage}
\end{figure}

\begin{figure}[h!]
    \centering
    \begin{minipage}{\textwidth}
    \subfigure{\includegraphics[width=\linewidth]{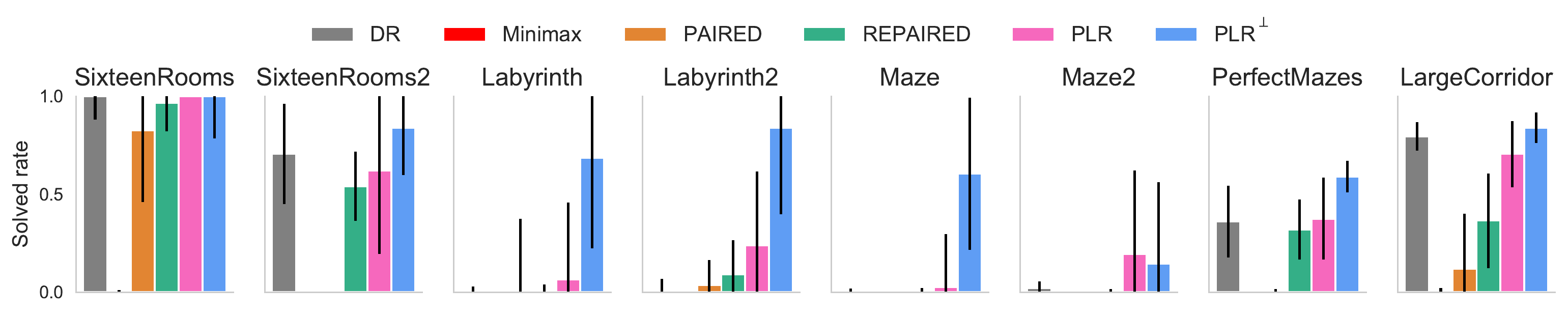}}
    \caption{\small{Zero-shot test performance on OOD environments when trained with a 25-block budget. The plots report the median and interquartile range of solved rates over 10 runs.}}
    \label{figure:mg25blocks_eval_250M}
    \end{minipage}
\end{figure}

We report the results of evaluating policies produced by each method after 250M training steps on each of the zero-shot transfer environments in Figure \ref{figure:mg25blocks_eval_250M} and Table \ref{table:minigrid_eval}. Examples of each test environment are presented in Figure \ref{figure:mg_test_mazes}. All replay-based UED methods lead to policies with statistically significantly ($p < 0.05$) higher test performance than PAIRED, and \plrabbrev{}, after 500M training steps, similarly improves over PLR when trained for an equivalent number of gradient updates (as replay rate is set to $0.5$). Note that for PAIRED and REPAIRED, we evaluate the protagonist policy. 

\begin{figure}[h!]
    \begin{minipage}{1\textwidth}
    \centering\subfigure{\includegraphics[width=\linewidth]{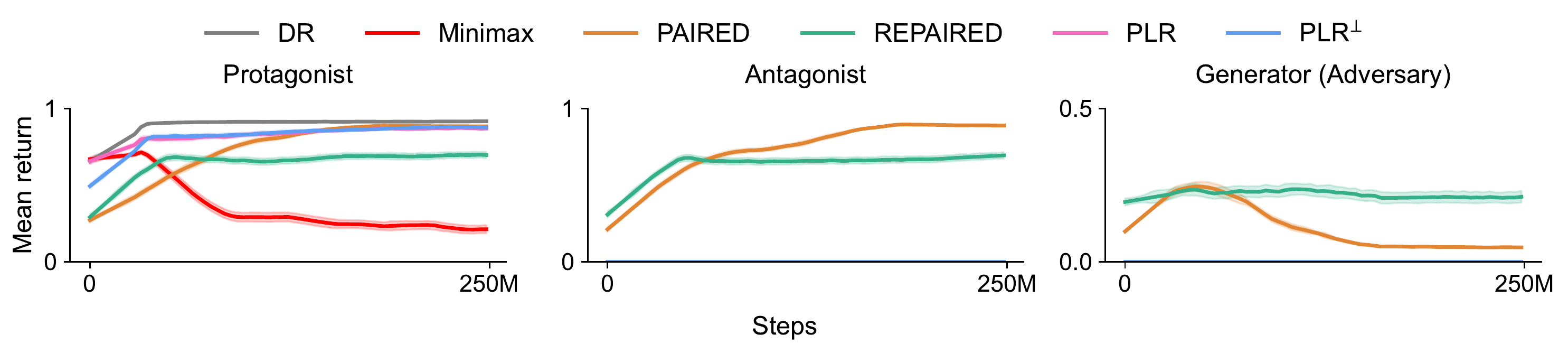}}
    \caption{\small{Training returns for each participating agent in each method, when trained with a 25-block budget. Plots show the mean and standard error over 10 runs.}}
    \label{figure:minigrid_training_returns_250M}
    \end{minipage}
\end{figure}

To provide a further sense of the training dynamics, we present the per-agent training returns for each method in Figure \ref{figure:minigrid_training_returns_250M}. Notably PAIRED results in antagonists that attain higher returns than the protagonist as expected. This dynamic takes on a mild oscillation, visible in the training return curve of the generator (adversary). As the protagonist adapts to the adversarial levels, the generator's return reduces, until the generator discovers new configurations that better exploit the relative differences between the two student policies. Notably, the adversary under REPAIRED seems to propose more difficult levels for both the protagonist and antagonist, while the resulting protagonist policy exhibits improved test performance, as seen in Figure \ref{figure:minigrid_curves}.

\begin{table}[!htb]
      \caption{\small{Mean test returns and standard errors on zero-shot transfer mazes for each method using a 25-block budget. Results are aggregated over 100 attempts for each maze across 10 runs per method. Bolded figures overlap in standard error with the method attaining the maximum mean test return in each row.}}
      \centering
      \scalebox{0.85}{
        \begin{tabular}{lllllllr}
\toprule
Environment &DR &Minimax &PAIRED &REPAIRED &PLR & \plrabbrev{} & \plrabbrev{} (500M) \\
\midrule
Labyrinth&$0.2\pm0.1$&$0.0\pm0.0$&$0.3\pm0.1$&$0.1\pm0.0$&$0.3\pm0.1$&$\mathbf{0.5\pm0.1}$&$\mathbf{0.7\pm0.1}$\\
Labyrinth2&$0.2\pm0.1$&$0.0\pm0.0$&$0.2\pm0.1$&$0.2\pm0.1$&$0.4\pm0.1$&$\mathbf{0.6\pm0.1}$&$\mathbf{0.8\pm0.1}$\\
LargeCorridor&$\mathbf{0.7\pm0.1}$&$0.1\pm0.1$&$0.3\pm0.1$&$0.5\pm0.1$&$\mathbf{0.7\pm0.1}$&$\mathbf{0.8\pm0.1}$&$\mathbf{0.8\pm0.1}$\\
Maze&$0.0\pm0.0$&$0.0\pm0.0$&$0.0\pm0.0$&$0.2\pm0.1$&$0.3\pm0.1$&$\mathbf{0.6\pm0.1}$&$\mathbf{0.5\pm0.1}$\\
Maze2&$0.0\pm0.0$&$0.0\pm0.0$&$0.1\pm0.1$&$0.1\pm0.1$&$\mathbf{0.4\pm0.1}$&$\mathbf{0.4\pm0.1}$&$\mathbf{0.5\pm0.1}$\\
PerfectMaze&$0.3\pm0.1$&$0.0\pm0.0$&$0.0\pm0.0$&$0.4\pm0.1$&$0.4\pm0.1$&$\mathbf{0.6\pm0.1}$&$\mathbf{0.5\pm0.1}$\\
SixteenRooms&$0.9\pm0.0$&$0.1\pm0.1$&$0.7\pm0.1$&$0.9\pm0.1$&$\mathbf{1.0\pm0.0}$&$0.8\pm0.1$&$\mathbf{1.0\pm0.0}$\\
SixteenRooms2&$\mathbf{0.7\pm0.1}$&$0.0\pm0.0$&$0.0\pm0.0$&$\mathbf{0.6\pm0.1}$&$\mathbf{0.5\pm0.1}$&$\mathbf{0.7\pm0.1}$&$\mathbf{0.7\pm0.1}$\\
\midrule
Mean &$0.4\pm0.0$&$0.0\pm0.0$&$0.2\pm0.0$&$0.4\pm0.0$&$0.5\pm0.1$&$\mathbf{0.6\pm0.1}$&$\mathbf{0.7\pm0.1}$\\
        \bottomrule
        \end{tabular}}
    \label{table:minigrid_eval}
\end{table}

\begin{figure}[h!]
    \begin{minipage}{1\textwidth}
    \centering\subfigure{\includegraphics[width=\linewidth]{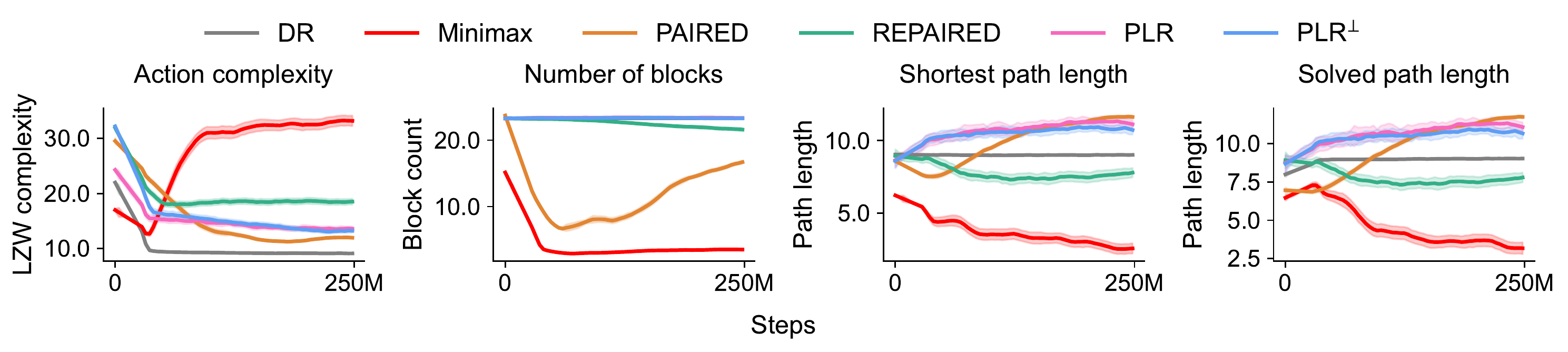}}
    \caption{\small{Complexity metrics of environments generated by the teacher throughout training with a 25-block budget. Plots show the mean and standard error of 10 runs.}}
    \label{figure:minigrid_complexity_all_250M}
    \end{minipage}
\end{figure}

Additional complexity metrics tracked during training are shown in Figure \ref{figure:minigrid_complexity_all_250M}. Alongside the number of blocks and shortest path length of levels seen during training, we also track solved path length and action complexity. Solved path length corresponds to the shortest path length from start position to goal in the levels successfully solved by the primary student agent (e.g. the protagonist in PAIRED). Action complexity corresponds to the Lempel-Ziv-Welch (LZW) complexity---a commonly used measure of string compressibility---of the action sequence taken during the primary student agent's trajectories. As expected, DR results in constant complexity for number of blocks and path length metrics. REPAIRED generates mazes with significantly greater complexity in terms of block count. The lower path lengths seen by REPAIRED suggest that it trains agents that more readily generalize to different path lengths, thereby pressuring the adversary to raise complexity in terms of block count. Further, given the high replay rates used, the REPAIRED adversary sees far fewer gradient updates with which to adjust its policy. As its shortest path lengths exceed that of PAIRED after adjusting proportionately by replay rate, foreseeably, over a longer period, the shortest path lengths generated by REPAIRED may meet or exceed that of PAIRED. In all cases, the action complexity reduces as the agent becomes more decisive, and we see that both PAIRED and REPAIRED lead to more decisive policies---as indicated by the simultaneously lower action complexity and greater level complexity in terms of higher block count (relative to DR) and, in the case of PAIRED, higher path length metrics. Lastly, it is interesting to note that while the random generator used by PLR produces levels of average complexity, the complexity of curated levels, as revealed in Figure \ref{figure:minigrid_curves}, is significantly higher and, in the case of path length, steadily increasing.

\label{appendix:subsubsec:minigrid_50b}
\subsubsection{Mazes with a 50-block budget}

\begin{figure}[h]
    \centering
    \begin{minipage}{\textwidth}
    \subfigure{\includegraphics[width=\linewidth]{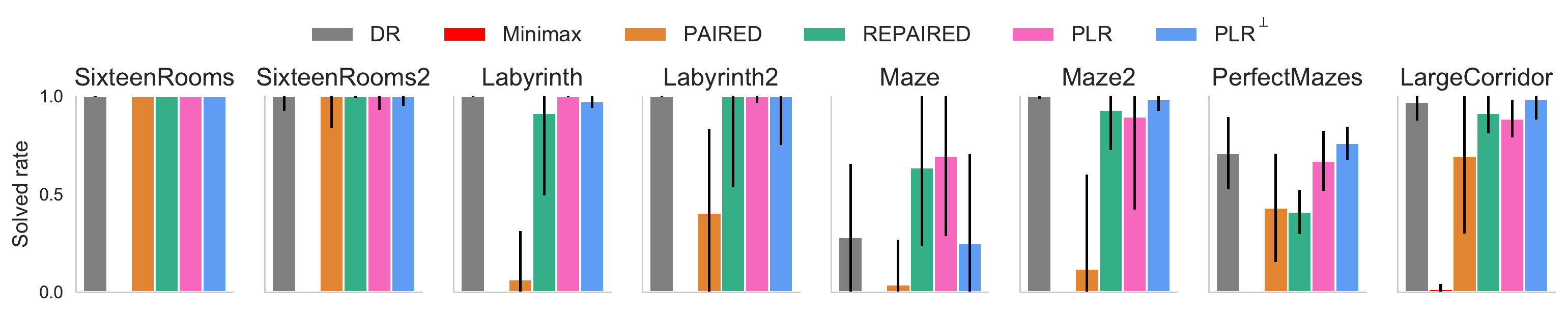}}
    \caption{\small{Zero-shot test performance on OOD environments when trained with a 50-block budget. The plots show the median and interquartile range of solved rates over 10 runs.}}
    \label{figure:mg50blocks_eval_500M}
    \end{minipage}
\end{figure}

\begin{figure}[h]
    \begin{minipage}{1\textwidth}
    \centering\subfigure{\includegraphics[width=\linewidth]{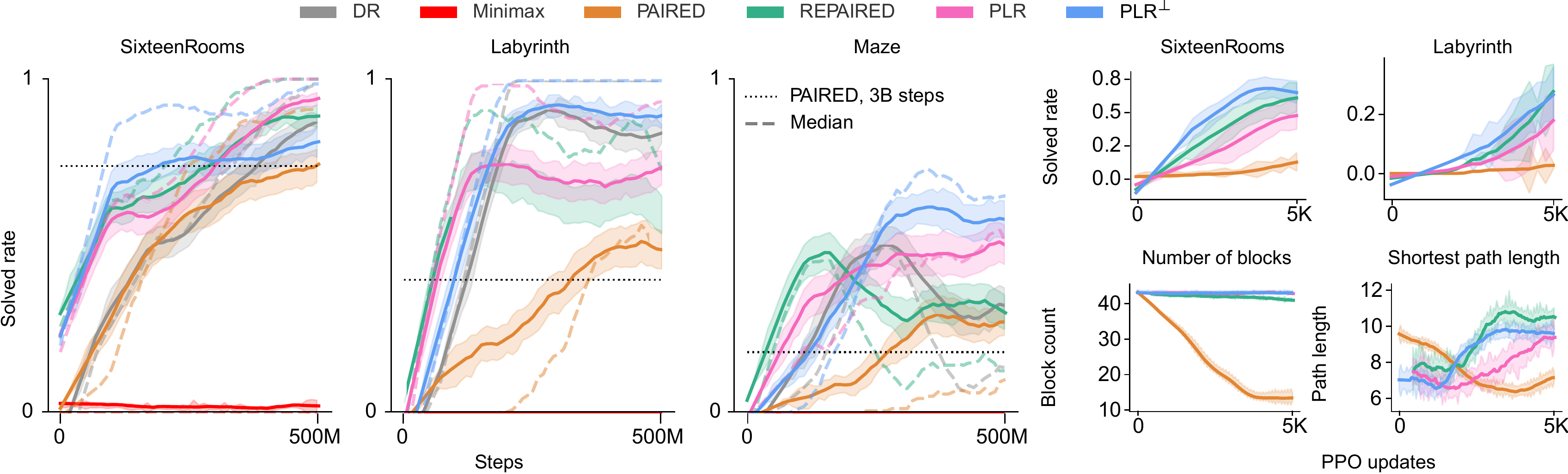}}
    \caption{\small{Test performance as a function of number of training steps with a 50-block budget (left), and test performance and complexity metrics as a function of number of PPO updates (right). The plots show the mean and standard error over 10 runs.}}
    \label{figure:mg50blocks_test_curves_500M}
    \end{minipage}
\end{figure}

\begin{figure}[h]
    \begin{minipage}{1\textwidth}
    \centering\subfigure{\includegraphics[width=\linewidth]{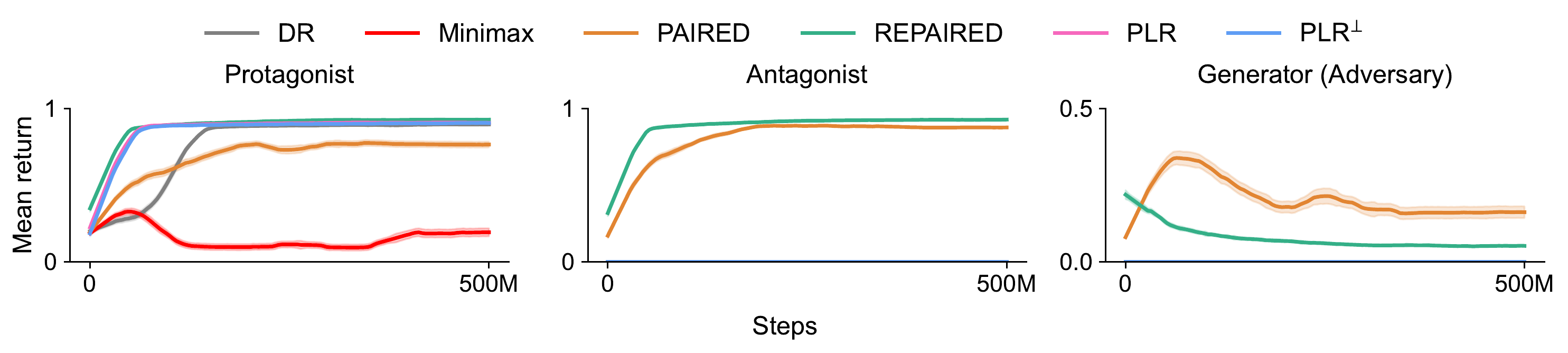}}
    \caption{\small{Training returns for each participating agent in each method when training with a 50-block budget. Plots show the mean and standard error over 10 runs.}}
    \label{figure:mg50blocks_agent_returns_500M}
    \end{minipage}
\end{figure}

\begin{figure}[h]
    \begin{minipage}{1\textwidth}
    \centering\subfigure{\includegraphics[width=\linewidth]{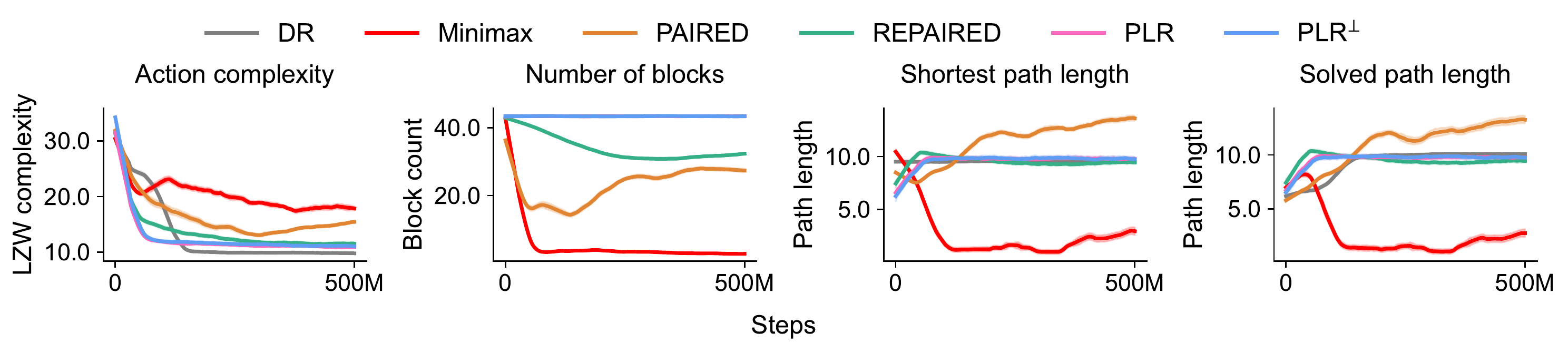}}
    \caption{\small{Complexity metrics of environments generated by the teacher throughout training with a 50-block budget. Plots show the mean and standard error of 10 runs.}}
    \label{figure:mg50blocks_complexity_500M}
    \end{minipage}
\end{figure}

\begin{figure}[h]
    \begin{minipage}{1\textwidth}
    \centering\subfigure{\includegraphics[width=0.65\linewidth]{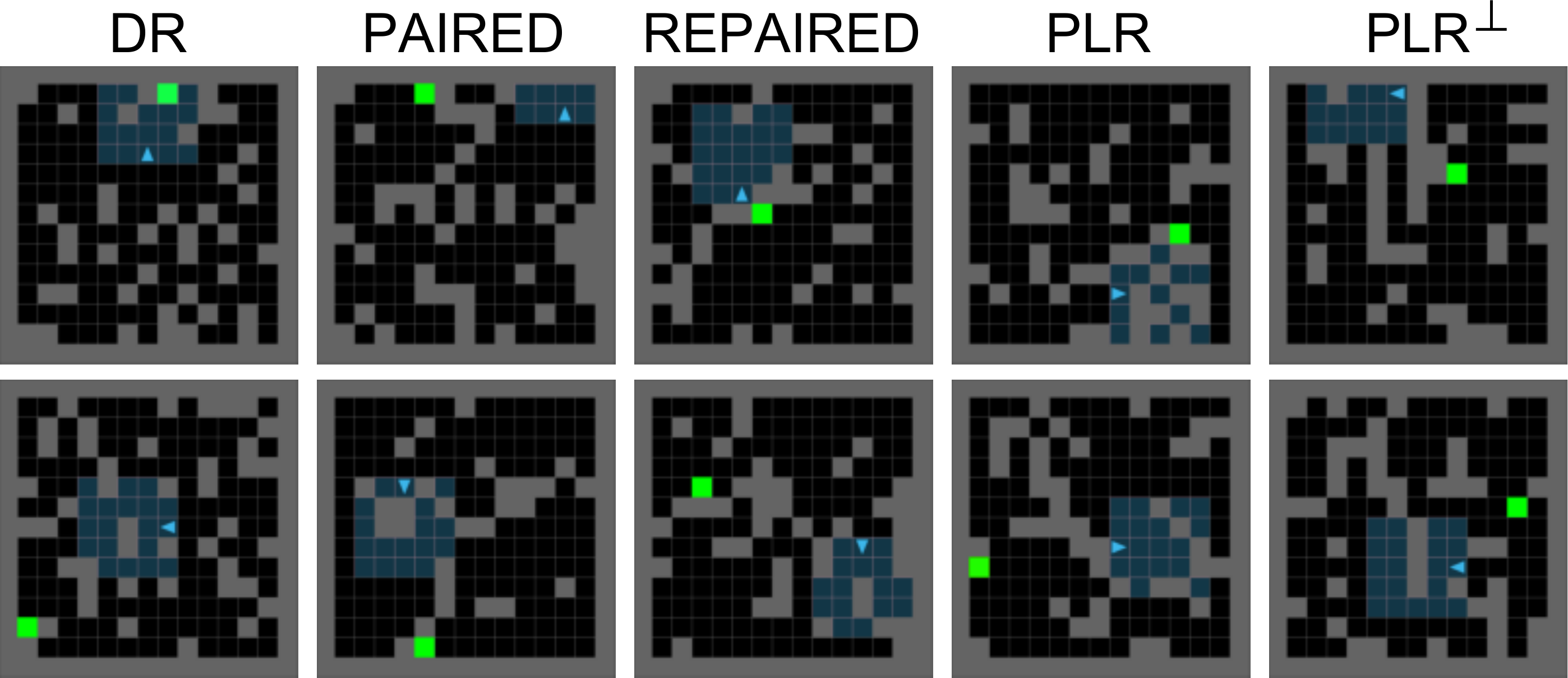}}
    \caption{\small{Example mazes generated by each method when using a 50-block budget.}}
    \label{figure:mg_50blocks_examples}
    \end{minipage}
\end{figure}

Similarly, Figures \ref{figure:mg50blocks_test_curves_500M}, \ref{figure:mg50blocks_agent_returns_500M}, and \ref{figure:mg50blocks_complexity_500M} report the training dynamics and test performance of agents trained using each method with a 50-block budget for 500M steps. Figure \ref{figure:mg50blocks_eval_500M} shows that DR and all replay-based methods are able to reach near perfect solve rates on most test mazes after 500M steps of training, with the exception of the Maze and PerfectMaze environments, where the test performances across methods are not markedly dissimilar, making the setting with a 50-block budget uninformative for assessing performance differences among these methods. The example mazes generated by each method, presented in Figure \ref{figure:mg_50blocks_examples}, shows that the larger block budget allows DR to sample mazes with greater structural complexity, leading to robust policies and diminishing the benefits of the UED methods studied. Therefore, in this work, we focus the main results for the maze domain on the more challenging setting with a 25-block budget. Note that the impact of the block budget on test performance further highlights the importance of properly adapting the training distribution for producing policies exhibiting high generality---a problem that our replay-based UED methods effectively address, as demonstrated by the results for the 25-block setting.

\pagebreak

\subsection{Extended Results for CarRacing}
\label{appendix:subsec:carracing}

\begin{figure}[t!]
    \begin{minipage}{\textwidth}
    \centering\subfigure{\includegraphics[width=\linewidth]{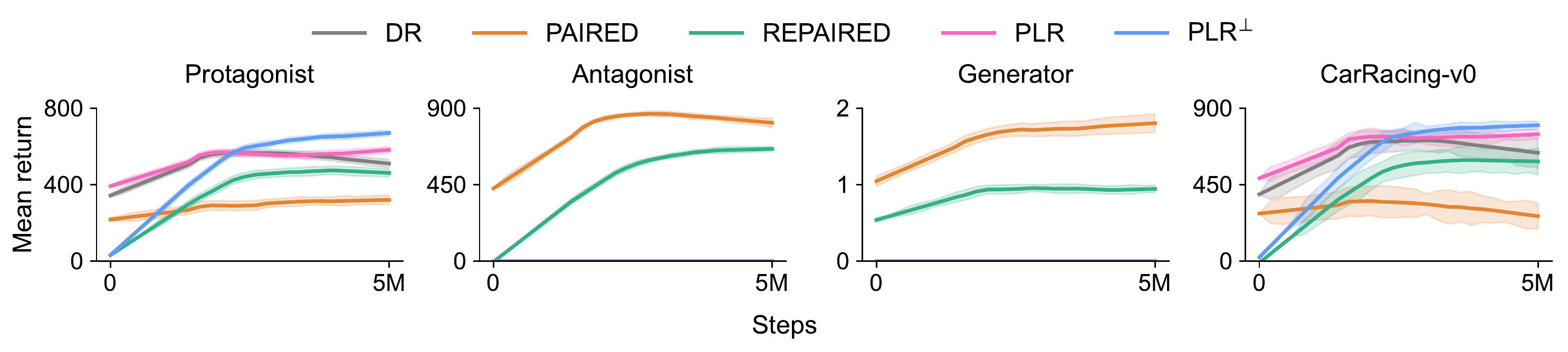}}
    \caption{\small{From left to right: Returns attained by the protagonist, antagonist, and generator (adversary) throughout training; the protagonist's zero-shot transfer performance on the original CarRacing-v0 during training. The mean and standard error over 10 runs are shown.}}
    \label{figure:carracing_training_returns_250M}
    \end{minipage}
\end{figure}

The training return plots for each agent, shown in Figure \ref{figure:carracing_training_returns_250M}, reveal that PAIRED's generator (adversary) overexploits the relative advantages of the antagonist over the protagonist, leading to a highly suboptimal protagonist policy. In fact, as shown in the right-most plot of Figure \ref{figure:carracing_training_returns_250M}, the resulting protagonist policies suffer such performance degradation from the adversarial curriculum that they can no longer even successfully drive on the original, simpler CarRacing tracks.

Additionally, we present per-track zero-shot transfer returns for the entire CarRacing-F1 benchmark after 5M training steps (equivalent to 40M environment interaction steps due to the usage of action repeat) in Table \ref{table:carracing_f1_benchmark}. Results report the mean and standard deviation over 100 attempts per track across 10 seeds. While DR acts as a strong baseline in terms of zero-shot generalization in this setting, \plrabbrev{} either attains the highest mean return, or matches the method achieving the highest return within standard error on all tracks. The mean performance of \plrabbrev{} across the full benchmark is statistically significantly higher ($p < 0.001$) than that of all other methods. Notably, PAIRED sees poor results, likely due to the generator's ability to overexploit the differences between antagonist and protagonist to detrimental effect in this domain. We see that REPAIRED mitigates this effect to a degree, resulting in more competitive policies. Note that due to the high compute overhead of training the AttentionAgent (8.2 billion steps of training over a population 256 agents) \citep{attentionagent}, we resorted to evaluating its mean F1 performance using the pre-trained model weights provided by the authors with their public code release. As a result, we only have a single training run for AttentionAgent. This means we cannot reliably compute standard errors for this baseline, but we believe that showing the performance for a single training seed of AttentionAgent on the F1 benchmark alongside our methods, as done in Figure \ref{figure:carracing_main_results}, nonetheless provides a useful comparison for further contextualizing the efficacy of our methods. This comparison highlights how, by only modifying the training curriculum, our methods produce policies with test returns exceeding that of AttentionAgent---which in contrast, uses a powerful attention-based policy and a much larger number of training steps.

\begin{sidewaysfigure}[h!]
    \begin{minipage}{1\textwidth}
    \centering\subfigure{\includegraphics[width=\linewidth]{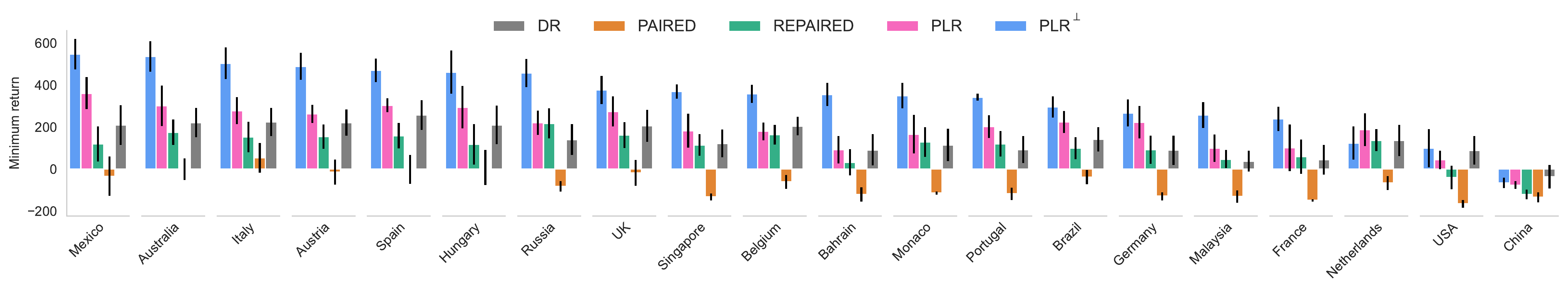}}
    \caption{\small{Minimum returns attained across 10 test episodes per track per seed. Bars report mean and standard error over 10 training runs.}}
    \label{figure:carracing_min_returns}
    \end{minipage}
\end{sidewaysfigure}

As a further analysis of robustness, we inspect the minimum returns over 10 attempts per track, averaged over 10 runs per method. We present these results (mean and standard error) in Figure \ref{figure:carracing_min_returns}. \plrabbrev{} achieves consistently higher minimum returns on average for many of the tracks compared to the other methods, including on the challenging Russia and USA tracks. The fact that simply curating random levels, as done by \plrabbrev{}, more reliably approaches a minimax regret policy than PAIRED and REPAIRED suggests that RL may not be an effective means for optimizing the PAIRED teacher.

\begin{table}[!htb]
      \caption{\small{
Mean test returns and standard errors of each method on the full F1 benchmark. Results are aggregated over 10 attempts for each track across 10 runs per method. Bolded figures overlap in standard error with the method attaining the maximum mean test return in each row. We see that \plrabbrev{} consistently either outperforms the other methods or matches PLR, the next best performing method. Note that we separately report the results of a single run for AttentionAgent due to its high compute overhead.
      }}
      \centering
      \scalebox{0.87}{
        \begin{tabular}{*{6}l||r}
\toprule
Track &DR &PAIRED &REPAIRED &PLR &PLR$^{\bot}$ & AttentionAgent \\
\midrule
Australia&$484\pm29$&$100\pm22$&$414\pm27$&$545\pm23$&$\mathbf{692\pm15}$ &826\\
Austria&$409\pm21$&$92\pm24$&$345\pm19$&$442\pm18$&$\mathbf{615\pm13}$ &511\\
Bahrain&$298\pm27$&$-35\pm19$&$295\pm23$&$411\pm22$&$\mathbf{590\pm15}$ &372\\
Belgium&$328\pm16$&$72\pm20$&$293\pm19$&$327\pm15$&$\mathbf{474\pm12}$ &668\\
Brazil&$309\pm23$&$76\pm18$&$256\pm19$&$387\pm17$&$\mathbf{455\pm13}$ &145\\
China&$115\pm24$&$-101\pm9$&$7\pm18$&$84\pm20$&$\mathbf{228\pm24}$& 344\\
France&$279\pm32$&$-81\pm13$&$240\pm29$&$290\pm35$&$\mathbf{478\pm22}$&153\\
Germany&$274\pm23$&$-33\pm16$&$272\pm22$&$388\pm20$&$\mathbf{499\pm18}$& 214\\
Hungary&$465\pm32$&$98\pm29$&$414\pm29$&$533\pm26$&$\mathbf{708\pm17}$& 769\\
Italy&$461\pm27$&$132\pm24$&$371\pm25$&$588\pm20$&$\mathbf{625\pm12}$&798\\
Malaysia&$236\pm25$&$-26\pm17$&$200\pm17$&$283\pm20$&$\mathbf{400\pm18}$ & 300\\
Mexico&$458\pm33$&$67\pm31$&$415\pm30$&$561\pm21$&$\mathbf{712\pm12}$&580\\
Monaco&$268\pm28$&$-28\pm18$&$256\pm26$&$360\pm32$&$\mathbf{486\pm19}$ &835\\
Netherlands&$328\pm26$&$70\pm20$&$307\pm21$&$\mathbf{418\pm21}$&$\mathbf{419\pm25}$& 131\\
Portugal&$324\pm27$&$-49\pm13$&$265\pm21$&$407\pm15$&$\mathbf{483\pm13}$& 606\\
Russia&$382\pm30$&$51\pm21$&$419\pm25$&$479\pm24$&$\mathbf{649\pm14}$ & 732\\
Singapore&$336\pm29$&$-35\pm14$&$274\pm21$&$386\pm22$&$\mathbf{566\pm15}$& 276\\
Spain&$433\pm24$&$134\pm24$&$358\pm24$&$482\pm17$&$\mathbf{622\pm14}$&759\\
UK&$393\pm28$&$138\pm25$&$380\pm22$&$456\pm16$&$\mathbf{538\pm17}$&729\\
USA&$263\pm31$&$-119\pm11$&$120\pm25$&$243\pm28$&$\mathbf{381\pm33}$& -192\\
\midrule
Mean&$341\pm22$&$19\pm15$&$293\pm18$&$408\pm12$&$\mathbf{534\pm7}$ & 477 \\
\bottomrule
        \end{tabular}}
    \label{table:carracing_f1_benchmark}
\end{table}

\clearpage

\begin{figure}[htb!]
    \centering\subfigure[DR]{\includegraphics[width=.185\linewidth]{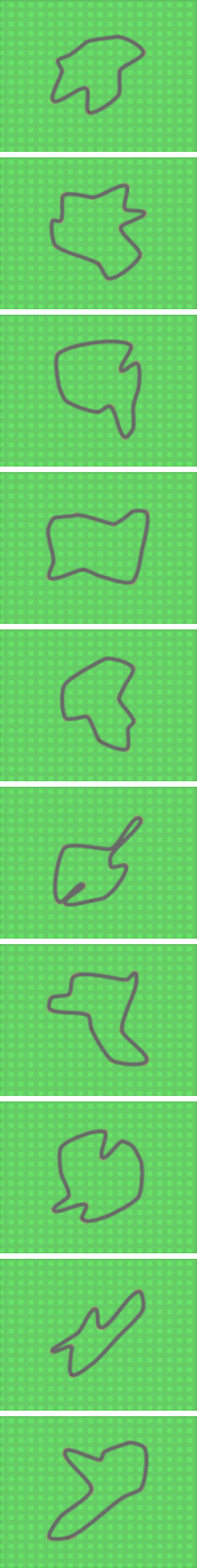}}
    \centering\subfigure[PAIRED]{\includegraphics[width=.185\linewidth]{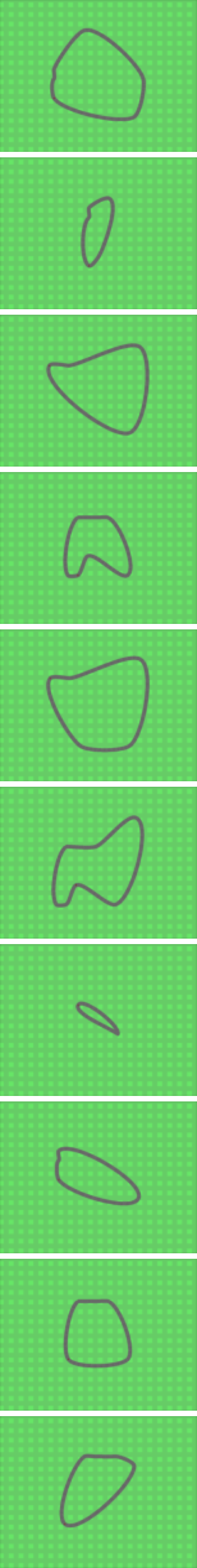}}
    \centering\subfigure[REPAIRED]{\includegraphics[width=.185\linewidth]{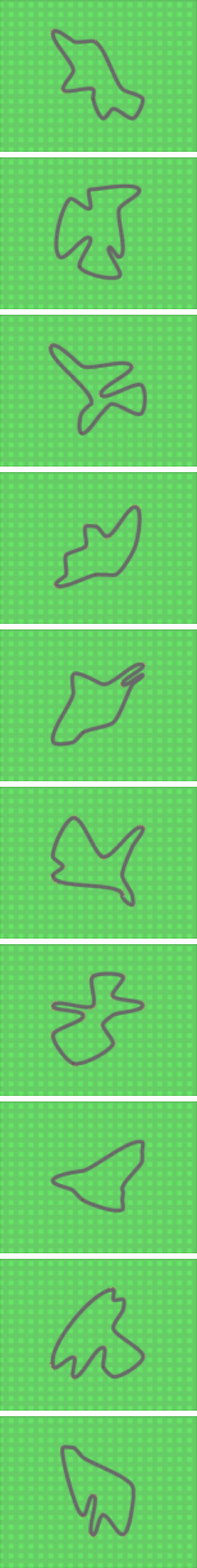}}
    \centering\subfigure[PLR]{\includegraphics[width=.185\linewidth]{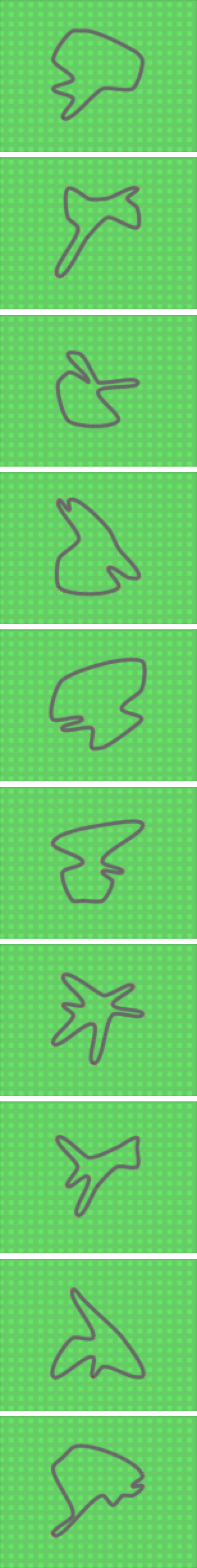}}
    \centering\subfigure[PLR$^{\perp}$]{\includegraphics[width=.185\linewidth]{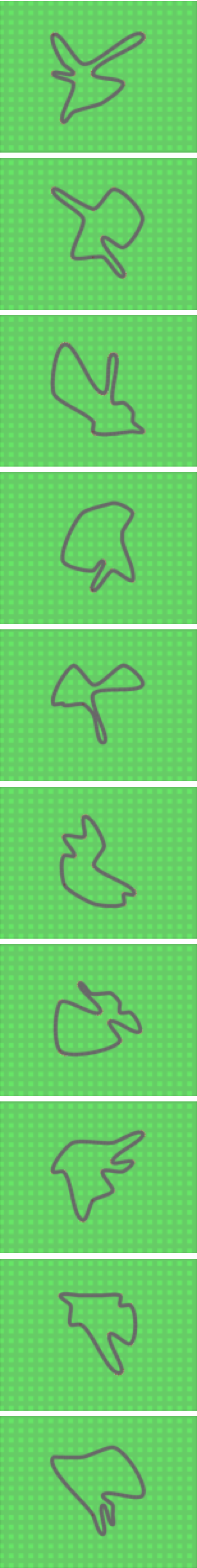}}
    
    \caption{\small{A randomly-selected set of CarRacing tracks generated by each method. (a)~Domain Randomization (DR) produces tracks of average complexity, with few sharp turns. (b)~PAIRED often overexploits the difference in the students, leading to simple tracks that incidentally favor the antagonist. (c)~REPAIRED mitigates this degeneracy, recovering track complexity. (d)~PLR and (e)~PLR$^\bot$ similarly generate tracks of considerable complexity, by prioritizing the most challenging randomly generated tracks.}}
    \label{figure:carracing_pics_appendix}
\end{figure}
\section{Experiment Details and Hyperparameters}
\label{appendix:exp_details}
\begin{table}[t!]
\caption{Hyperparameters used for training each method in the maze and car racing environments.}
\label{table:hyperparams}
\begin{center}
\scalebox{0.87}{
\begin{tabular}{lrr}
\toprule
\textbf{Parameter} & MiniGrid & CarRacing \\
\midrule
\emph{PPO} & & \\
$\gamma$ & 0.995 & 0.99 \\
$\lambda_{\text{GAE}}$ & 0.95 & 0.9 \\
PPO rollout length & 256 & 125 \\
PPO epochs & 5 & 8 \\
PPO minibatches per epoch & 1 & 4 \\
PPO clip range & 0.2 & 0.2 \\
PPO number of workers & 32 & 16 \\
Adam learning rate & 1e-4 & 3e-4 \\
Adam $\epsilon$ & 1e-5 & 1e-5 \\
PPO max gradient norm & 0.5 & 0.5 \\
PPO value clipping & yes & no \\
Return normalization & no & yes \\
Value loss coefficient & 0.5 & 0.5 \\
Student entropy coefficient & 0.0 & 0.0 \\

\addlinespace[10pt]
\emph{PLR} & & \\
Replay rate, $p$ & 0.5 & 0.5 \\
Buffer size, $K$ & 4000 & 8000 \\
Scoring function & MaxMC & positive value loss \\
Prioritization & rank & proportional \\
Temperature, $\beta$ & 0.3 & 1.0 \\
Staleness coefficient, $\rho$ & 0.3 & 0.7 \\

\addlinespace[10pt]
\emph{PLR}$^{\bot}$ & & \\
Scoring function & MaxMC & positive value loss \\

\addlinespace[10pt]
\emph{PAIRED} & & \\
Generator entropy coefficient & 0.0 & 0.0 \\

\addlinespace[10pt]
\emph{REPAIRED} & & \\
Generator entropy coefficient & 0.0 & 0.01 \\
Scoring function & MaxMC & MaxMC \\

\bottomrule 
\end{tabular}
}
\end{center}
\end{table}

This section details the environments, agent architectures, and training procedures used in our experiments discussed in Section \ref{sec:experiments}. We use PPO to train both student and generator policies in all experiments. Section \ref{sec:experiments} reports results for each method using the best hyperparameter settings, which we summarize in Figure \ref{table:hyperparams}. Note that unless specified, PPO hyperparameters are shared between student and teacher, and PLR hyperparameters are shared between \algoabbrev{} and REPAIRED. The procedures for determining the hyperparameter choices for each environment are detailed below, in Sections \ref{appendix:minigrid_details} and \ref{appendix:carracing_details}. 

\subsection{Partially-Observable Navigation (MiniGrid)}
\label{appendix:minigrid_details}

\textbf{Environment details} Our mazes are based on MiniGrid \citep{gym_minigrid}. Each maze consists of a $15 \times 15$ grid, where each cell can contain a wall, the goal, the agent, or navigable space. The student agent receives a reward of $1-T/T_{\textnormal{max}}$ upon reaching the goal, where $T$ is the episode length and $T_{\textnormal{max}}$ is the maximum episode length (set to 250). Otherwise, the agent receives a reward of 0 if it fails to reach the goal. The observation space consists of the agent's orientation (facing north, south, east, or west) and the $7\times7$ grid immediately in front of and including the agent. This grid takes the form of a 3-channel integer encoding. The action space consists of 7 total actions, though mazes only make use of the first three: turn left, turn right, and forward. We do not mask out irrelevant actions.

\textbf{Level generation} Each maze is fully surrounded by walls, resulting in $13\times13=169$ cells in which the generator can place walls, the goal, and the agent. Starting from an initially empty maze (except the bordering walls), the generator is given a budget of $W=50$ steps in which it can choose a grid cell in which to place a wall. Placing a wall in a cell already containing a wall results in a no-opt. After wall placement, the generator then chooses cells for the goal and the agent's starting position. If either of these cells collides with an existing wall, a random empty cell is chosen. At each time step, the generator teacher receives the full grid observation of the developing maze, the one-hot encoding of the current time step, as well as a 50-dimensional random noise vector, where each component is uniformly sampled from $[0.0, 1.0]$.

\textbf{Generator architecture} We base the generator architecture on the the original model used for the PAIRED adversary in \citep{paired}. This model encodes the full grid observation using a convolution layer ($3\times 3$ kernel, stride length $1$, 128 filters) followed by a ReLU activation layer over the flattened convolution outputs. The current time step is embedded into a 10-dimensional space, which is concatenated to the grid embedding, along with the random noise vector. This combined representation is then passed through an LSTM with hidden dimension 256, followed by two fully-connected layers, each with a hidden dimension 32 and ReLU activations, to produce the action logits over the 169 possible cell choices. We further ablated the LSTM and found that its absence preserves the performance of the minimax generator in both 25-block and 50-block settings, as well as that of the PAIRED generator in the 50-block setting, as expected given that the full grid and time step form a Markov state. However, the PAIRED generator struggles to learn without an LSTM in the 25-block setting. We believe PAIRED's improved performance when using an LSTM-based generator in the 25-block setting is due to the additional network capacity provided by the LSTM. Therefore, in favor of less compute time, our experiments only used an LSTM-based generator for PAIRED in the 25-block setting.

\textbf{Student architecture} The student policy architecture resembles the LSTM-based generator architecture, except the student model uses a convolution with 16 filters to embed its partial observation; does not use a random noise vector; and instead of embedding the time step, embeds the student's current direction into a 5-dimensional latent space.

\textbf{Choice of hyperparameters} We base our choice of hyperparameters for student agents and generator (i.e. the adversary) on \citep{paired}. We also performed a coarse grid search over the student entropy coefficient in $\{0.0, 0.01\}$, generator entropy coefficient in $\{0.0, 0.005, 0.01\}$, and number of PPO epochs in $\{5, 20\}$ for both students and generator, as well as the choice of including an LSTM in the student and generator policies. We selected the best performing settings based on average return on the validation levels of SixteenRooms, Labyrinth, and Maze over 3 seeds. Our final choices are summarized in \ref{table:hyperparams}. The main deviations from the settings in \citep{paired} are the choice of removing the generator's LSTM (except for PAIRED with 25 blocks) and using fewer PPO epochs (5 instead of 20). For PLR, we searched over replay rate, $p$, in $\{0.5, 0.95\}$ and level buffer size, $K$, in $\{500, 2000, 4000\}$, temperature $\beta$ in $\{0.1, 0.3\}$, and choice of scoring function in $\{\textnormal{positive value loss}, \textnormal{MaxMC}\}$. The final PLR hyperparameter selection was then also used for \plrabbrev{} and REPAIRED, except for the scoring function, over which we conducted a separate search for each method.

\textbf{Zero-shot levels} We make use of the challenging test mazes in \citep{paired}: SixteenRooms, requiring navigation through up to 16 rooms to find a goal; Labyrinth, requiring traversal of a spiral labyrinth; and Maze, requiring the agent to find a goal in a binary-tree maze, which requires the agent to successfully backtrack from dead ends. To more comprehensively test the agent's zero-shot transfer performance on OOD classes of mazes, we introduce Labyrinth2, a rotated version of Labyrinth; Maze2, another variant of a binary-tree maze; PerfectMaze, a procedurally-generated maze environment; and LargeCorridor, another procedurally-generated maze environment, where the goal position is randomly chosen to lie at the end of one of the corridors, thereby testing the agent's ability to perform backtracking. Figure \ref{figure:main_minigrid} provides screenshots of these mazes.

\textbf{Compute} All maze-navigating agents were trained using Tesla V100 GPUs. DR required approximately 40 hours to reach 250 million training steps; minimax, 50 hours; PLR variants, 100 hours; and PAIRED variants, 160 hours. In total, our main experimental results, across 25-block and 50-block runs, required roughly 18,300 hours (around 763 days) of training.

\subsection{CarRacing}
\label{appendix:carracing_details}

\textbf{Environment details} Each track consists of a closed loop around which the student agent must drive a full lap. In order to increase the expressiveness of the original CarRacing, we reparameterized the tracks using Bézier curves. In our experiments, each track consists of a Bézier curve \citep{bezier_ref} based on 12 randomly sampled control points within a fixed radius, $B/2$, of the center of the $B\times B$ playfield. 
The track consists of a sequence of $L$ polygons. When driving over each previously unvisited polygon, the agent receives a reward equal to $1000/L$. The student additionally receives a reward of -0.1 at each time step. Aligning with the methodology of \citep{carracing_ppo}, we do not penalize the agent for driving out of the playfield boundaries, terminate episodes if the agent drives too far off track, and repeat every selected action for 8 steps. The student observation space consists of a $96\times96\times3$ pixel observation with RGB channels with a clipped, egocentric, bird's-eye view of the vehicle centered horizontally in the top $84\times96$ portion of the frame. The remaining $12\times96$ portion of the frame consists of the dashboard visualizing the agent's latest action and return. Note that despite the lossiness of the downsampled dashboard, our hyperparameter sweep for the best PPO settings found that including the full frame enabled better performance. Given this observation, the student then decides on a 3-dimensional continuous action, where the components correspond to control values for steer (torque, in $[-1.0, 1.0]$), gas (acceleration, in $[0.0, 1.0]$), and brake (deceleration, in $[0.0, 1.0]$). 

\textbf{Level generation} Starting from an empty track, the adversary generates a sequence of 12 control points, one per time step, spaced within a fixed radius, $B/2$ of the center $O$ of the playfield. The agent always begins centered at the track polygon closest to $0^{\circ}$ relative to $O$, facing counterclockwise.

\textbf{Generator architecture} At each time step, the generator policy receives the set of all control points so far generated, the current time step encoded as a one-hot vector, and a 16-dimensional random noise vector. The control points are spatially encoded in a $10\times10$ grid, called the \emph{sketch}, representing a downsampled and discretized version of the playfield bounds within which the generated track resides. Choosing a control point then corresponds to selecting one of the cells in this grid. After the control points are chosen, each control point's cell coordinates are upscaled to match the  original playfield scale. This ensures no two control points are too close together, preventing areas of excessive track overlapping. The sketch is embedded using two $2\times2$ convolutions using a stride length of 1 with 8 and 16 channels respectively, each followed by a ReLU layer. The flattened outputs of this sequence of convolutions is then concatenated with an 8-dimensional embedding of the time step and the random noise vector. This combined embedding is then fed through two fully connected layers, each with a hidden size of 256, where the first is followed by a ReLU activation, to produce the policy logits over the 100 choices of control points. Note that we mask out any cells in the sketch that have already been chosen to prevent double selection of the same control point. We also experimented with outputing continuous, downsampled control points in $[0.0, 1.0]$ by learning the $\alpha$ and $\beta$ parameters of a $\textnormal{Beta}$ distribution for each of $x$ and $y$ coordinates instead of categorical logits, but found this latter parameterization led to slower learning of generator policies, where the generator policy tended to remain close to or revert to an approximately uniformly random policy.

\textbf{Student architecture} The student policy architecture is based on the competitive PPO implementation in \citep{carracing_ppo}, which was used as a baseline for AttentionAgent in \citep{attentionagent}. This architecture consists of an image embedding module composed of a stack of 2D convolutions with square kernels of sizes 2, 2, 2, 2, 3, 3, channel outputs of 8, 16, 32, 64, 128, 256, and stride lengths of 2, 2, 2, 2, 1, 1 respectively, resulting in a 256-dimensional image embedding. The image embedding is then passed through a fully connected layer with a hidden size of 100, followed by a ReLU layer. This latter output is then fed through two separate fully-connected layers, each with hidden size of 100 and output dimension equal to the action dimension, followed by softplus activations. We then add 1 to each component of these two output vectors, which serve as the $\alpha$ and $\beta$ parameters respectively for the $\textnormal{Beta}$ distributions used to sample each action dimension. When training the student, we normalize rewards by dividing rewards by the running standard deviation of returns so far encountered. 

\textbf{Choice of hyperparameters} To determine the best hyperparameters for the student agents, we performed a grid search, in which we trained a  student agent with domain randomization for 300 PPO updates. The grid search covered PPO learning rate in $\{0.001, 0.0003\}$, $\lambda_{\textnormal{GAE}}$ in $\{0.0, 0.5, 0.9\}$, number of PPO epochs in $\{4, 8\}$, PPO number of minibatches per epoch in $\{2, 4, 8\}$, value loss coefficient in $\{0.5, 2.0\}$, whether to grayscale frames, whether to crop frames (i.e remove the dashboard portion), and whether to normalize returns. Further, we found entropy regularization tended to hurt performance of the student policy. Similar to the sharing of PPO hyperparameters between student and generator in \citep{paired}, we then shared the best PPO hyperparameters for the student with the generator, with the exception of searching over separate choices for the entropy coefficient in $\{0.0, 0.01\}$. We selected the best performing settings based on average return on the validation levels of F1-Italy, F1-Singapore, and F1-Germany over 3 seeds. For PLR, we searched over replay rate, $p$, in $\{0.5, 0.95\}$, level buffer size $K$, in $\{500, 2000, 4000, 8000\}$, replay prioritization in $\{\text{rank}, \text{proportional}\}$, staleness coefficient $\rho$ in $\{0.3, 0.7\}$, and replay distribution temperature $\beta$ in \{0.1, 1.0, 2.0\}. The best settings for PLR were then shared with \algoabbrev{} and REPAIRED, except for the scoring function, over which we performed a separate search for each method. 

\begin{figure}[h!]
\vspace{-3mm}
    \begin{minipage}{1\textwidth}
    \centering\subfigure[Australia]{\includegraphics[width=.24\linewidth]{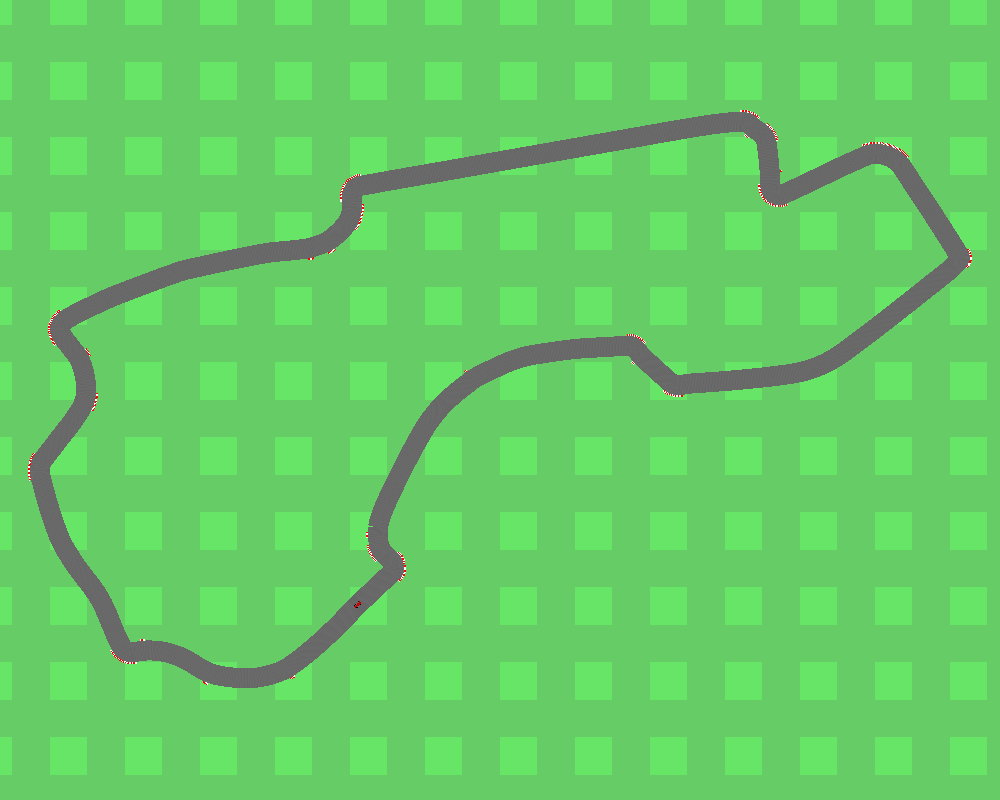}}
    \centering\subfigure[Austria]{\includegraphics[width=.24\linewidth]{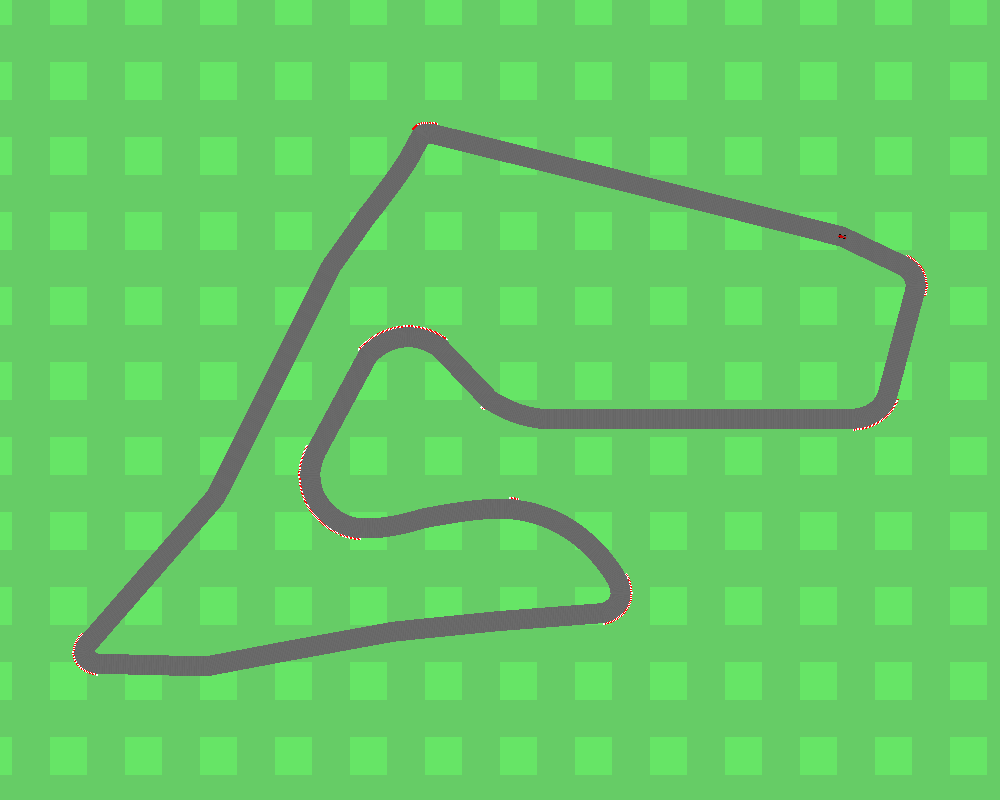}}  
    \centering\subfigure[Bahrain]{\includegraphics[width=.24\linewidth]{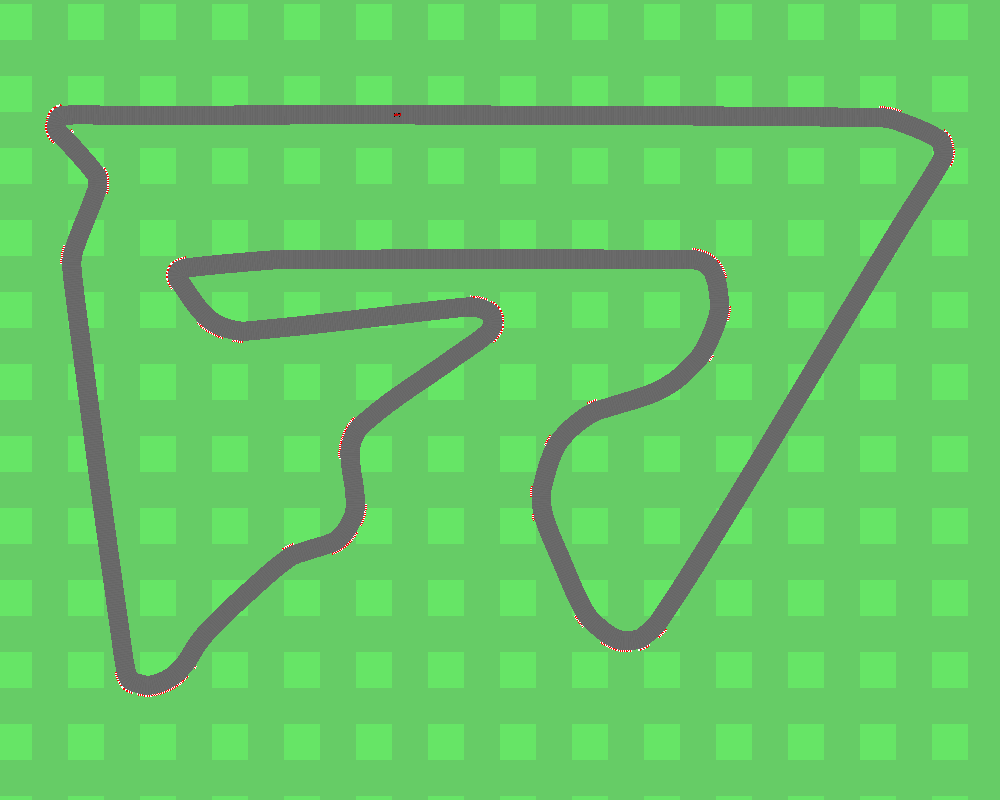}}  
    \centering\subfigure[Belgium]{\includegraphics[width=.24\linewidth]{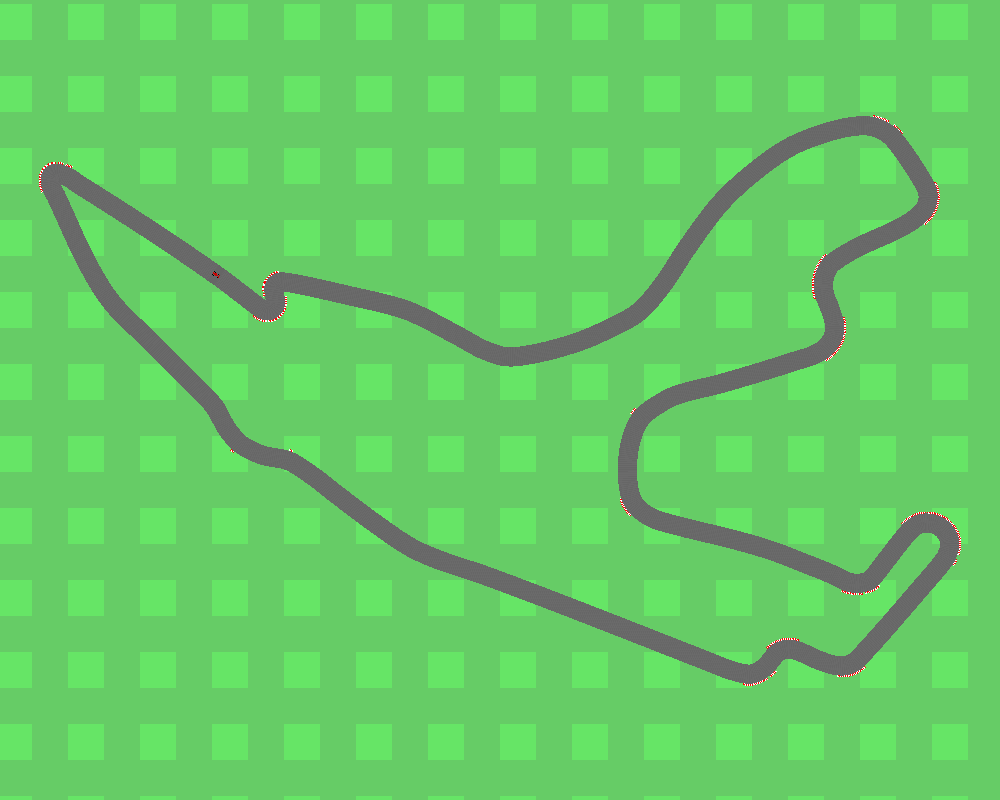}}
    \centering\subfigure[Brazil]{\includegraphics[width=.24\linewidth]{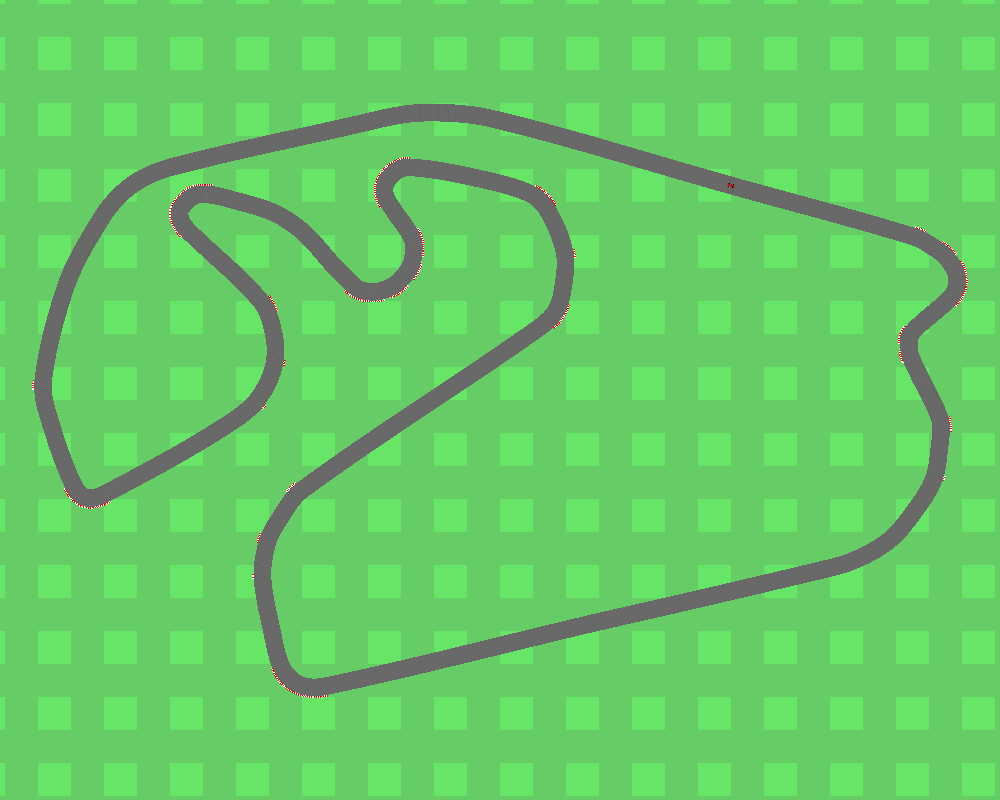}}
    \centering\subfigure[China]{\includegraphics[width=.24\linewidth]{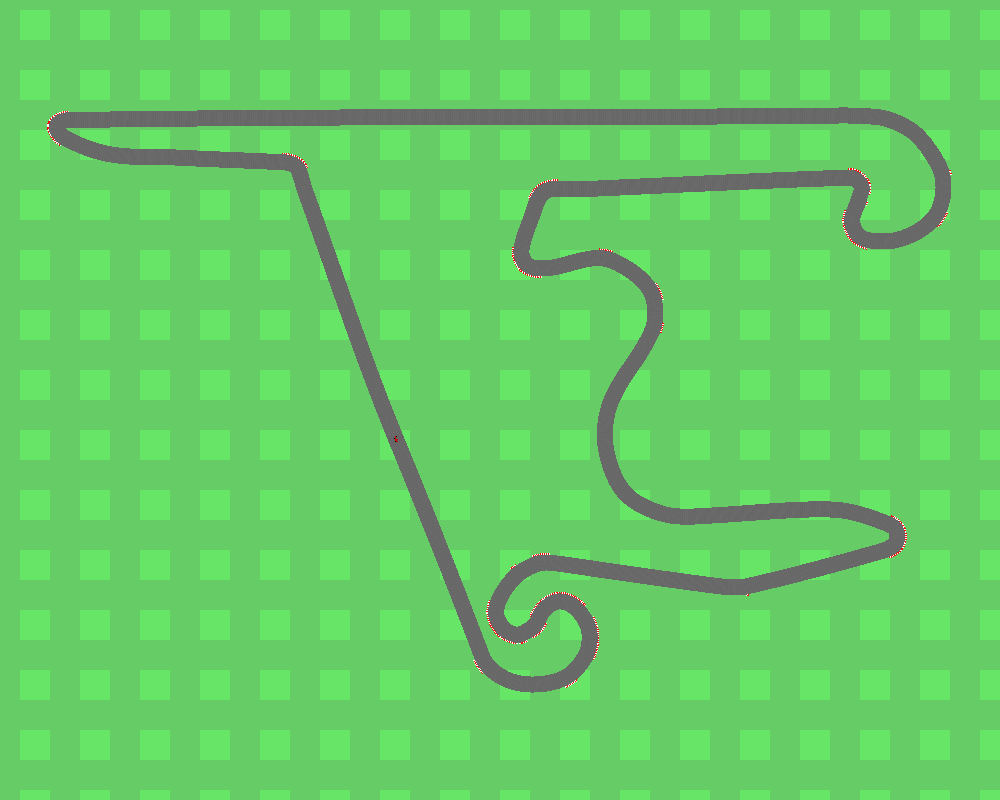}}
    \centering\subfigure[France]{\includegraphics[width=.24\linewidth]{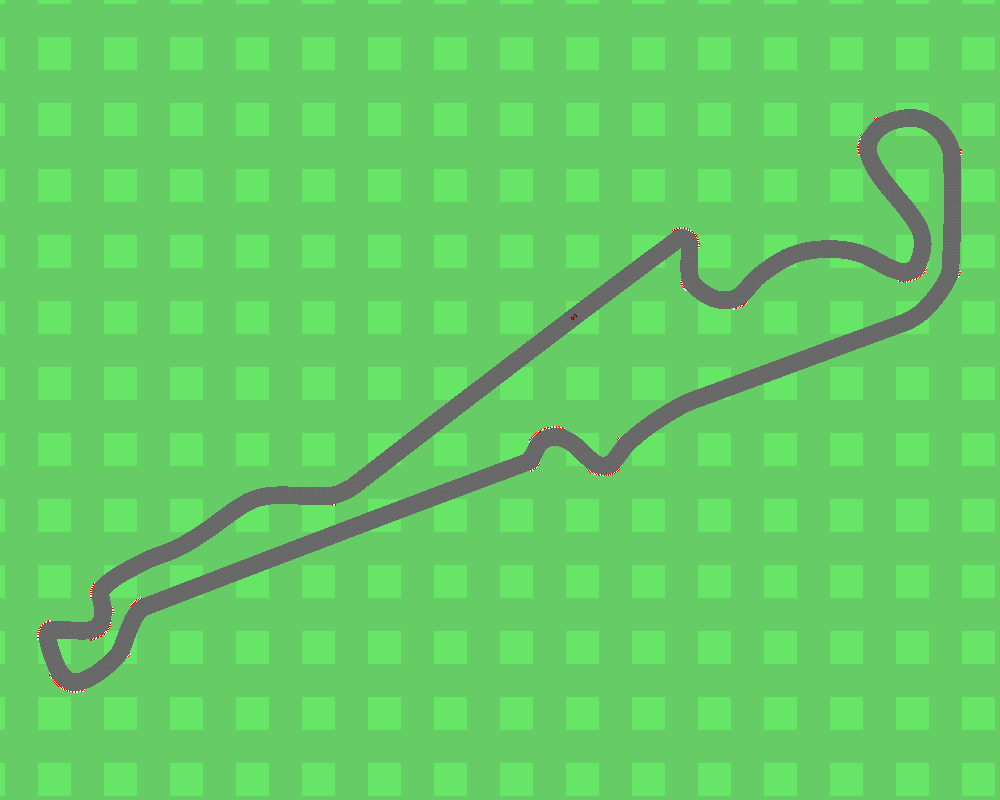}}
    \centering\subfigure[Germany]{\includegraphics[width=.24\linewidth]{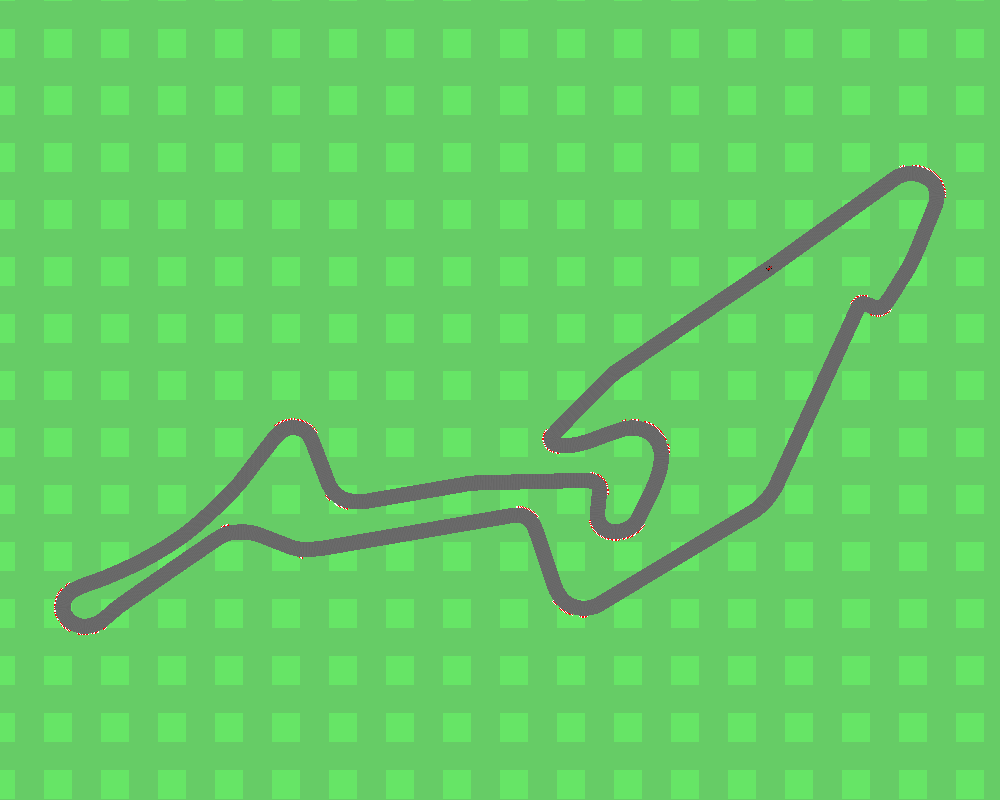}}  
    \centering\subfigure[Hungary]{\includegraphics[width=.24\linewidth]{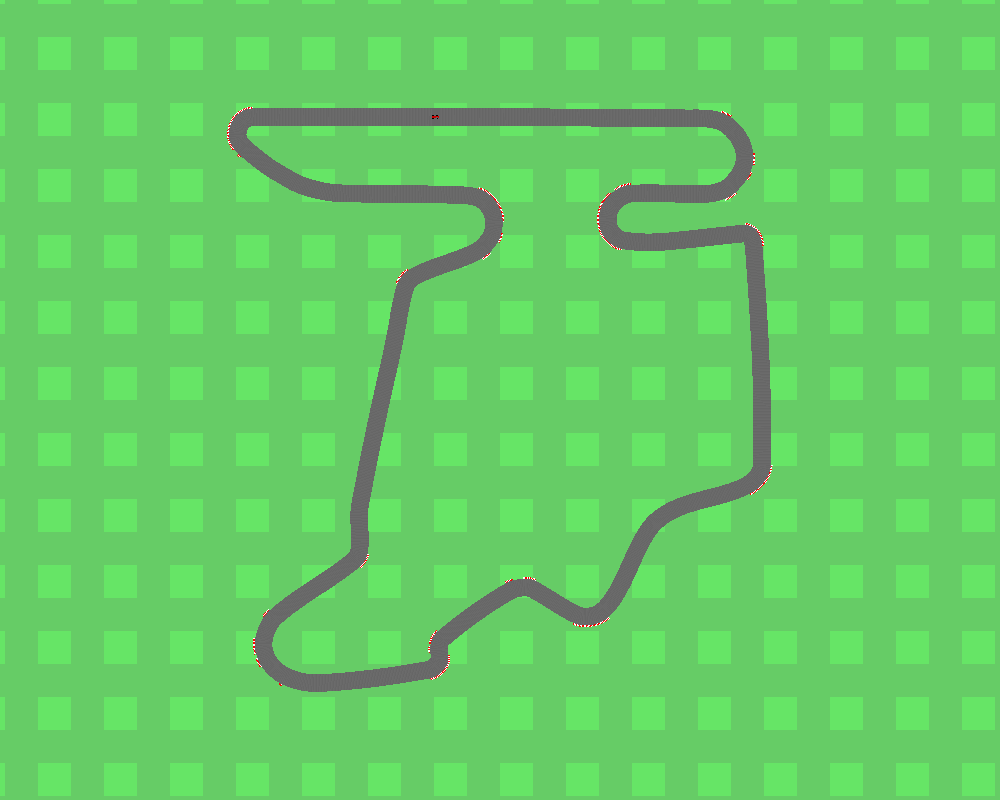}}  
    \centering\subfigure[Italy]{\includegraphics[width=.24\linewidth]{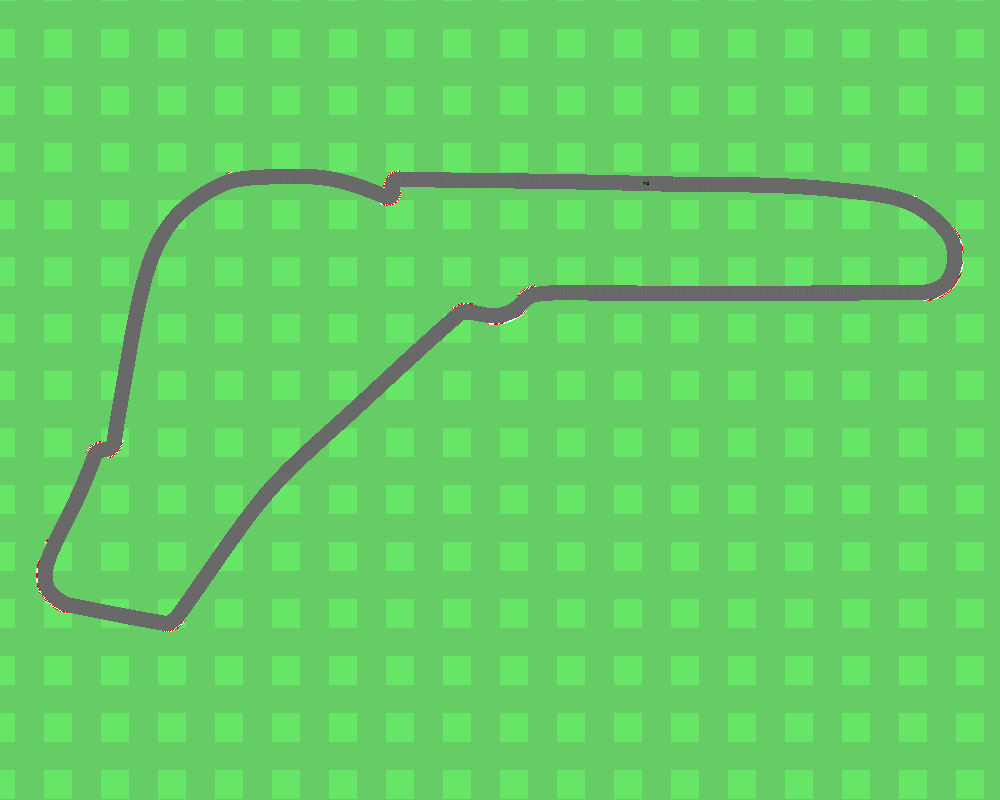}}
    \centering\subfigure[Malaysia]{\includegraphics[width=.24\linewidth]{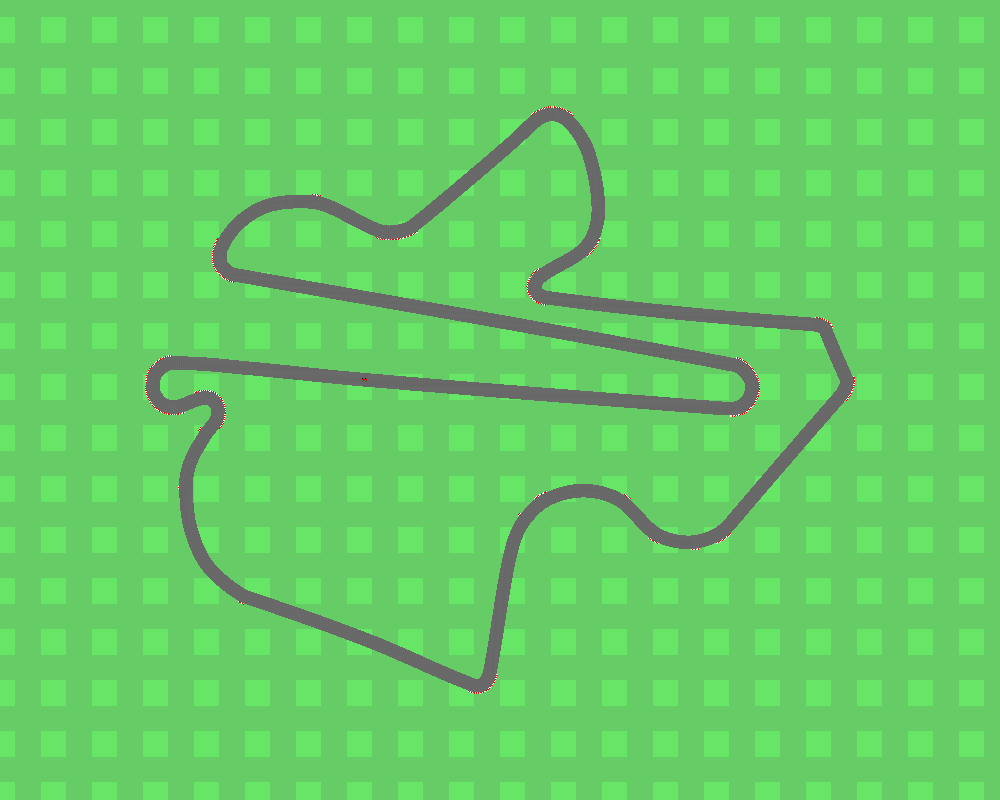}}
    \centering\subfigure[Monaco]{\includegraphics[width=.24\linewidth]{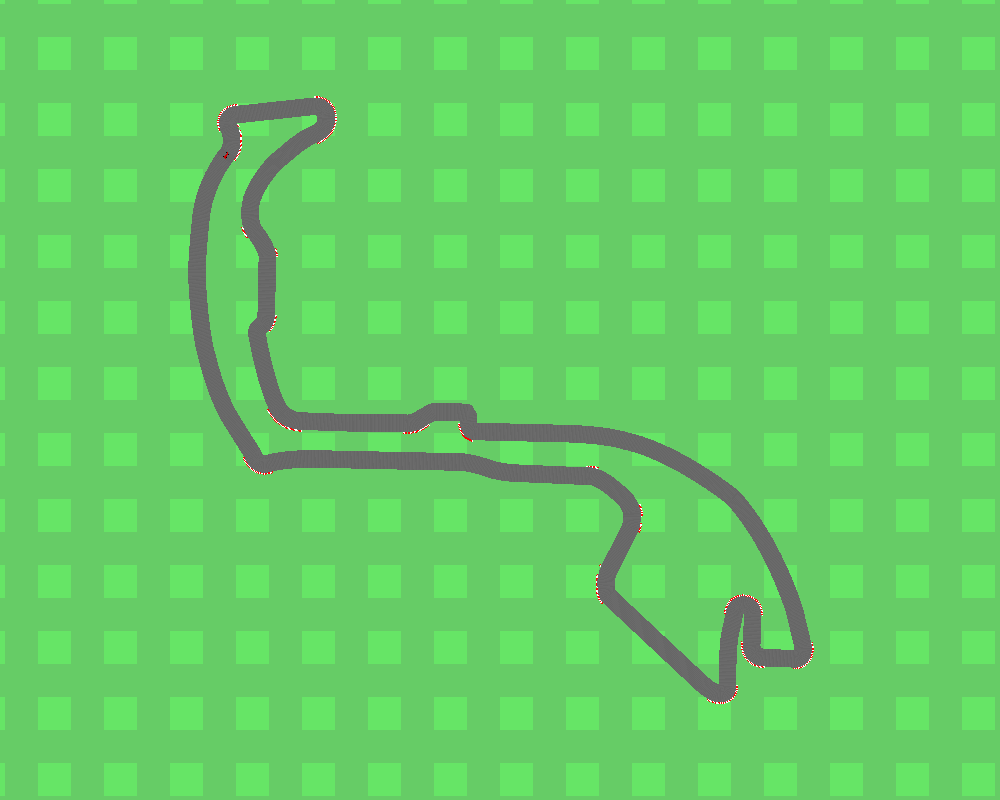}}
    \centering\subfigure[Mexico]{\includegraphics[width=.24\linewidth]{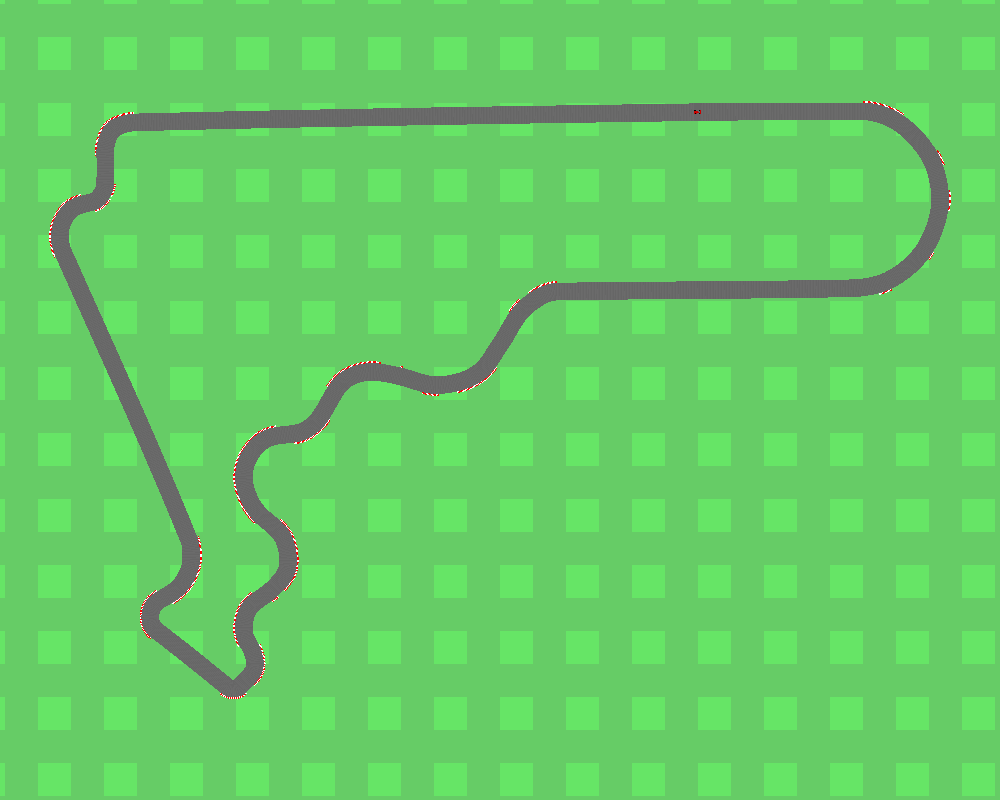}}
    \centering\subfigure[Netherlands]{\includegraphics[width=.24\linewidth]{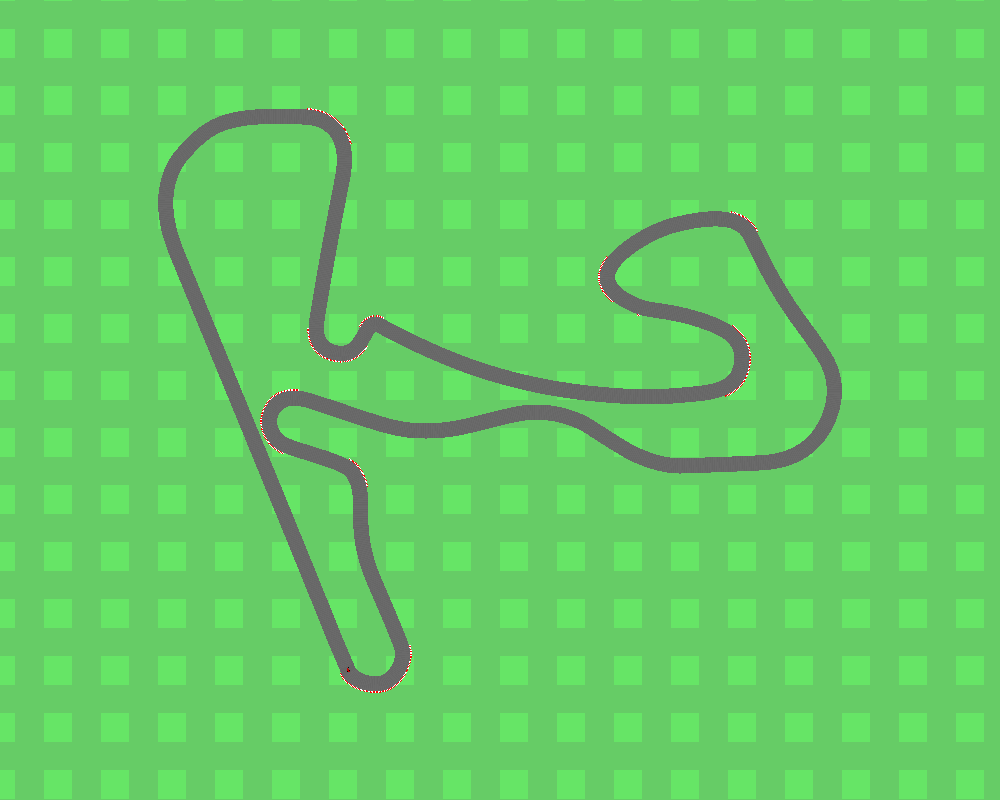}}  
    \centering\subfigure[Portugal]{\includegraphics[width=.24\linewidth]{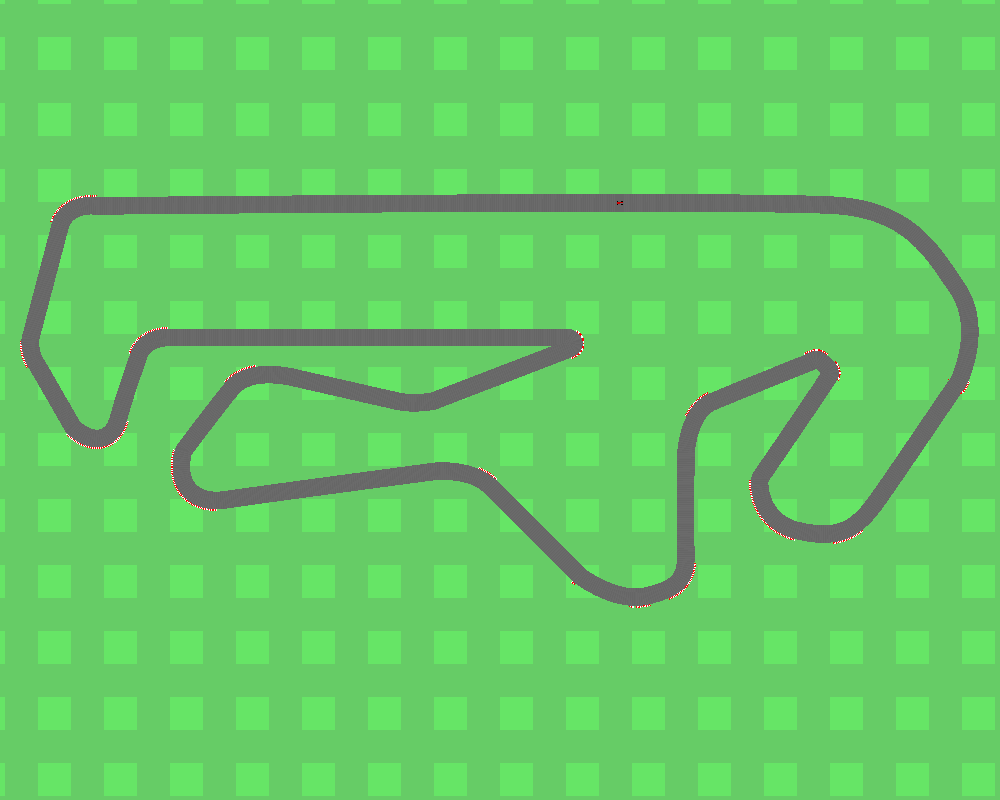}}  
    \centering\subfigure[Russia]{\includegraphics[width=.24\linewidth]{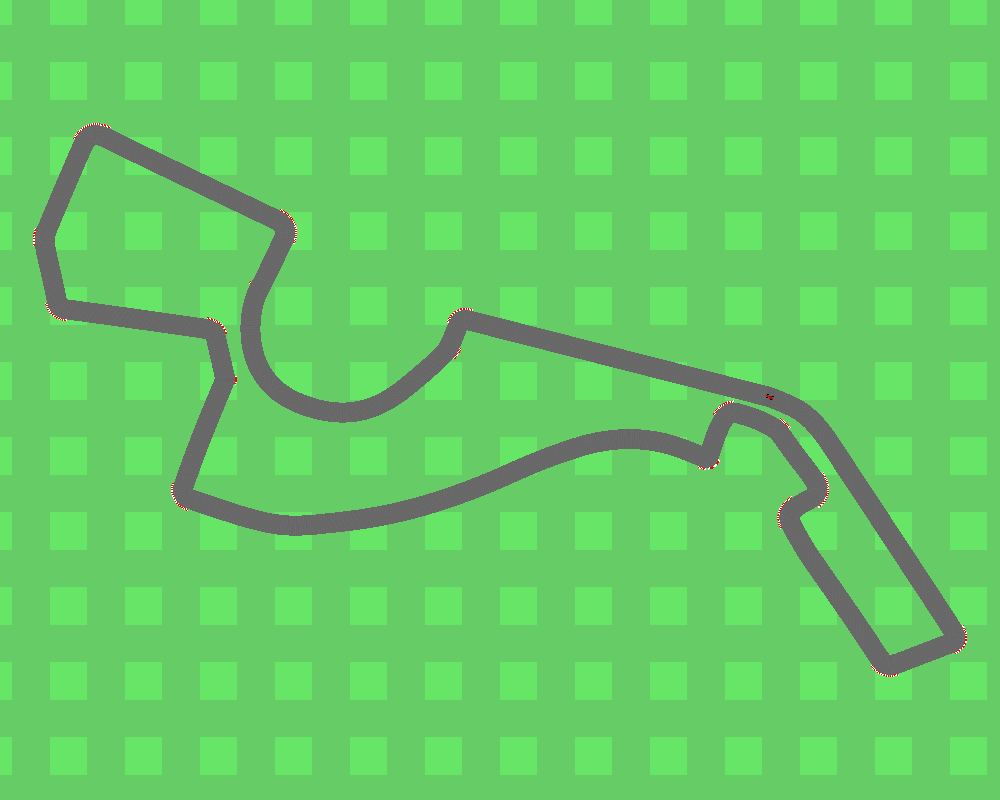}}
    \centering\subfigure[Singapore]{\includegraphics[width=.24\linewidth]{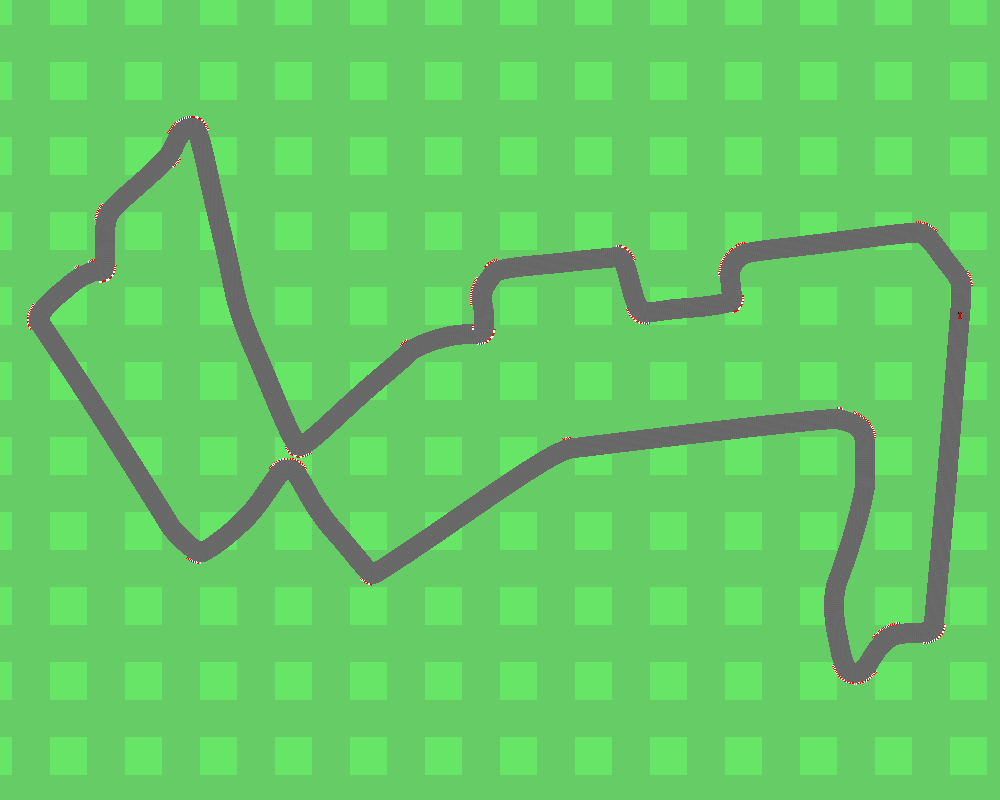}}
    \centering\subfigure[Spain]{\includegraphics[width=.24\linewidth]{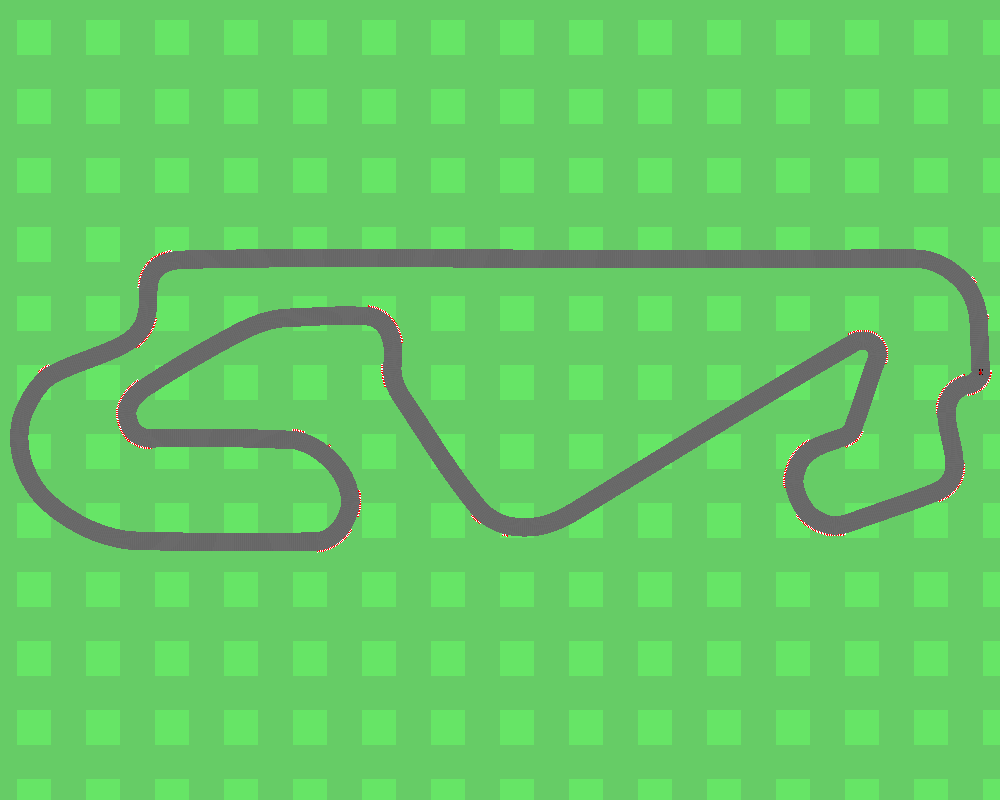}}
    \centering\subfigure[UK]{\includegraphics[width=.24\linewidth]{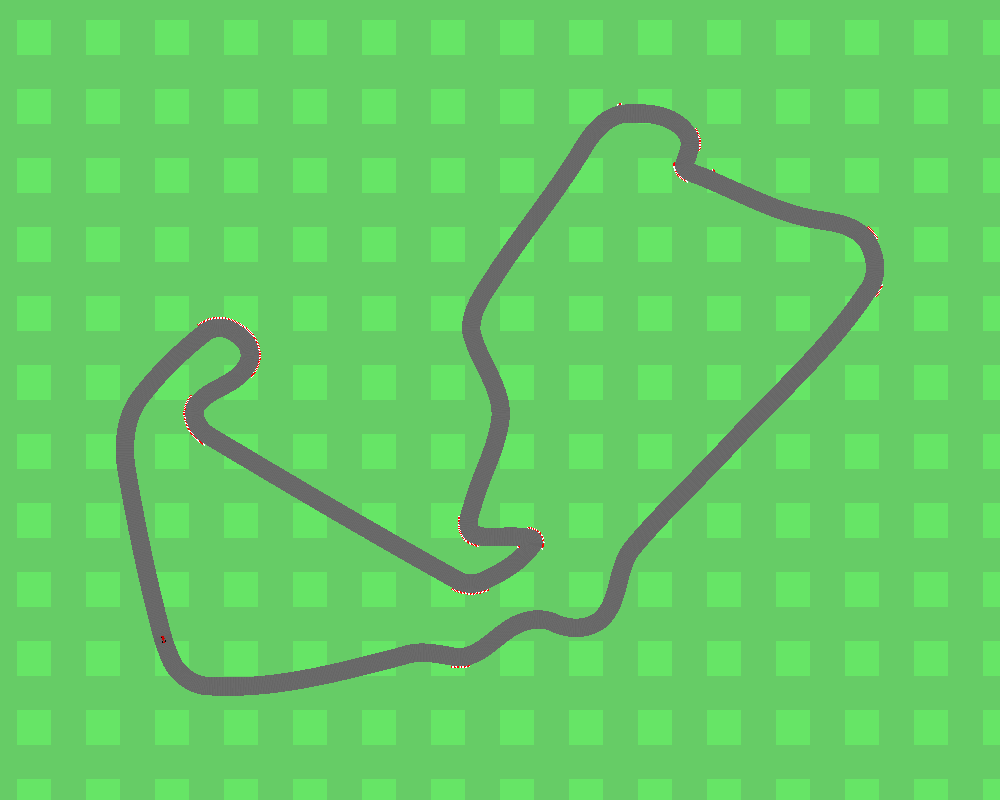}}
    \centering\subfigure[USA]{\includegraphics[width=.24\linewidth]{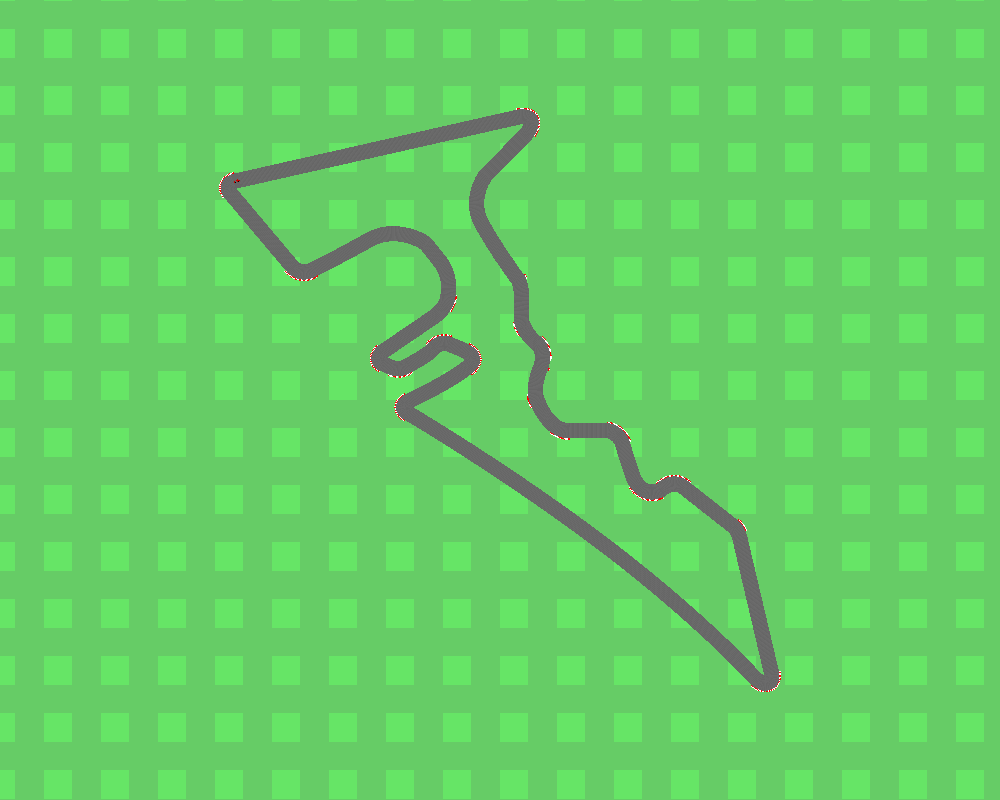}}  
    \caption{\small{All tracks in the CarRacing-F1 benchmark used for evaluating zero-shot generalization.}}
    \label{figure:f1_tracks}
    \end{minipage}
\end{figure}

\textbf{Zero-shot levels} Our zero-shot levels are based on 20 real-world Formula One (F1) tracks designed to challenge professional racecar drivers. We predominantly selected tracks based on recent F1 seasons, including some historical favorites such as the Nürburgring Grand Prix.\footnote{We chose not to include the Japanese and Canadian Grand Prix due to the overlapping tracks at Suzuka and the Circuit Gilles Villeneuve.} This collection of tracks, which we call CarRacing-F1, provides a new benchmark for testing robustness and zero-shot generalization in a continuous control setting. Importantly, these tracks are strictly out-of-distribution and of higher complexity with respect to the training levels, as they cannot be represented by Bézier curves limited to 12 control points. Moreover, each F1 track requires more time steps to solve (1500 or 2000) than allotted for the training tracks (1000). Table \ref{table:carracing_f1_tracks} provides per-track descriptions, and Figure \ref{figure:f1_tracks} shows bird's-eye views of each track.

\textbf{Compute} All car-racing agents were trained on Tesla V100 GPUs. DR and PLR variants required approximately 18 hours to reach 5 million training steps, while PAIRED variants, 24 hours. Our experiments in this domain entailed a total of roughly 9,600 hours (around 400 days) of training.
 
\begin{table}[!htb] 
      \caption{\small{Descriptions for each track in the CarRacing-F1 benchmark.}}
      \centering
      \scalebox{0.87}{
        \begin{tabular}{llr}
        \toprule
         Environment & Real-world track  & Max episode steps   \\
        \midrule
        Australia & Albert Park & 1500 \\
        Austria & Red Bull Ring & 1500 \\
        Bahrain & Bahrain International Circuit & 2500 \\
        Belgium &  Circuit de Spa-Francorchamps & 1500  \\
        Brazil &  Autódromo José Carlos Pace & 2000  \\
        China &  Shanghai International Circuit & 2500 \\
        France & Circuit Paul Ricard & 2000  \\
        Germany & Nürburgring & 2000  \\
        Hungary & Hungaroring & 2000  \\
        Italy & Monza Circuit & 1500  \\
        Malaysia & Sepang International Circuit  & 2500 \\
        Mexico & Autódromo Hermanos Rodríguez  & 2000  \\
        Monaco & Circuit de Monaco  & 1500  \\
        Netherlands & Circuit Zandvoort  & 2000 \\
        Portugal & Algarve International Circuit  & 2500  \\
        Russia &  Sochi Autodrom & 1500  \\
        Singapore &  Marina Bay Street Circuit & 2000  \\
        Spain &  Circuit de Barcelona-Catalunya & 2000  \\
        UK & Silverstone & 2000   \\
        USA & Circuit of the Americas, Austin & 2000  \\

        \bottomrule
        \end{tabular}}
    \label{table:carracing_f1_tracks}
\end{table}

\end{document}